\documentclass[journal,12pt,onecolumn]{IEEEoj}
\usepackage{cite}
\usepackage{amsmath,amssymb,amsfonts}
\usepackage{algorithmic}
\usepackage{graphicx,color}
\usepackage{textcomp}
\usepackage{xcolor}
\usepackage{hyperref}
\let\labelindent\relax
\usepackage{enumitem}
\usepackage{array}
\usepackage{multicol}
\usepackage{float}
\usepackage{dblfloatfix}
\usepackage[labelsep=period,bf]{caption}
\usepackage{subcaption}
\usepackage{parskip}
\usepackage{placeins}
\usepackage[export]{adjustbox}

\usepackage{booktabs}
\usepackage{colortbl}
\usepackage{diagbox}

\colorlet{tableheadcolor}{gray!25} 
\newcommand{\headcol}{\rowcolor{tableheadcolor}} %
\colorlet{tablerowcolor}{gray!10} 
\newcommand{\topline}{\arrayrulecolor{black}\specialrule{0.1em}{\abovetopsep}{0pt}%
	\arrayrulecolor{tableheadcolor}\specialrule{\belowrulesep}{0pt}{0pt}%
	\arrayrulecolor{black}}
\newcommand{\midline}{\arrayrulecolor{tableheadcolor}\specialrule{\aboverulesep}{0pt}{0pt}%
	\arrayrulecolor{black}\specialrule{\lightrulewidth}{0pt}{0pt}%
	\arrayrulecolor{white}\specialrule{\belowrulesep}{0pt}{0pt}%
	\arrayrulecolor{black}}

\newcommand{\bottomlinec}{\arrayrulecolor{tablerowcolor}\specialrule{\aboverulesep}{0pt}{0pt}%
	\arrayrulecolor{black}\specialrule{\heavyrulewidth}{0pt}{\belowbottomsep}}%

\def\OJlogo{\vspace{-4pt}\includegraphics[height=18pt]{new_logo}}

\def\authorrefmark#1{\ensuremath{^{\textbf{#1}}}}

\def\BibTeX{{\rm B\kern-.05em{\sc i\kern-.025em b}\kern-.08em
    T\kern-.1667em\lower.7ex\hbox{E}\kern-.125emX}}
\AtBeginDocument{\definecolor{ojcolor}{cmyk}{0.93,0.59,0.15,0.02}}
\def\OJlogo{\vspace{-4pt}\hskip-4pt\includegraphics[height=18pt]{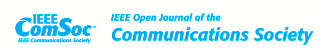}}

\usepackage{setspace}
\usepackage{geometry}
\begin{document}

\receiveddate{XX Month, XXXX}
\reviseddate{XX Month, XXXX}
\accepteddate{XX Month, XXXX}
\publisheddate{XX Month, XXXX}
\currentdate{18 May, 2024}
\doiinfo{OJCOMS.2024.011100}

\title{Early-Scheduled Handover Preparation in 5G NR Millimeter-Wave Systems}

\author{Dino Pjanić\authorrefmark{1,2}, Student Member, IEEE, Alexandros Sopasakis\authorrefmark{3}, Member, IEEE, \\ Andres Reial\authorrefmark{1}, Senior Member, IEEE and Fredrik Tufvesson\authorrefmark{2}, Fellow, IEEE}
\affil{Ericsson AB, Lund, Sweden}
\affil{Department of Electrical and Information Technology, Lund University, Lund, Sweden}
\affil{Department of Mathematics, Lund University, Lund, Sweden}
\corresp{Corresponding author: Dino Pjanić (email: dino.pjanic@ericsson.com, dino.pjanic@eit.lth.se).}
\authornote{``The work related to the contribution of D. P. and F. T. is partially sponsored by the Swedish Foundation for Strategic Research and Ericsson AB, Sweden. The work for A.S. is partially supported by grants from eSSENCE no. 138227, Vinnova 2020-033375, Formas 2022-00757 and Swedish National Space Board.
The training and data handling was enabled by resources provided by the Swedish National Infrastructure for Computing (SNIC), partially funded by the Swedish Research Council through grant agreement no. 2018-05973.''}
\markboth{Early-Scheduled Handover Preparation in 5G NR Millimeter-Wave Systems}{Pjanić \textit{et al.}}

\begin{abstract}
    The handover (HO) procedure is one of the most critical functions in a cellular network driven by measurements of the user channel of the serving and neighboring cells. The success rate of the entire HO procedure is significantly affected by the preparation stage. As massive Multiple-Input Multiple-Output (MIMO) systems with large antenna arrays allow resolving finer details of channel behavior, we investigate how machine learning can be applied to time series data of beam measurements in the Fifth Generation (5G) New Radio (NR) system to improve the HO procedure. This paper introduces the Early-Scheduled Handover Preparation scheme designed to enhance the robustness and efficiency of the HO procedure, particularly in scenarios involving high mobility and dense small cell deployments. Early-Scheduled Handover Preparation focuses on optimizing the timing of the HO preparation phase by leveraging machine learning techniques to predict the earliest possible trigger points for HO events. We identify a new early trigger for HO preparation and demonstrate how it can beneficially reduce the required time for HO execution reducing channel quality degradation. These insights enable a new HO preparation scheme that offers a novel, user-aware, and proactive HO decision making in MIMO scenarios incorporating mobility.
  
\end{abstract}


\maketitle

\section{INTRODUCTION}
\label{Intro} 
\begin{figure}[t!]
		\centering 
		\includegraphics[width=1.0\linewidth]{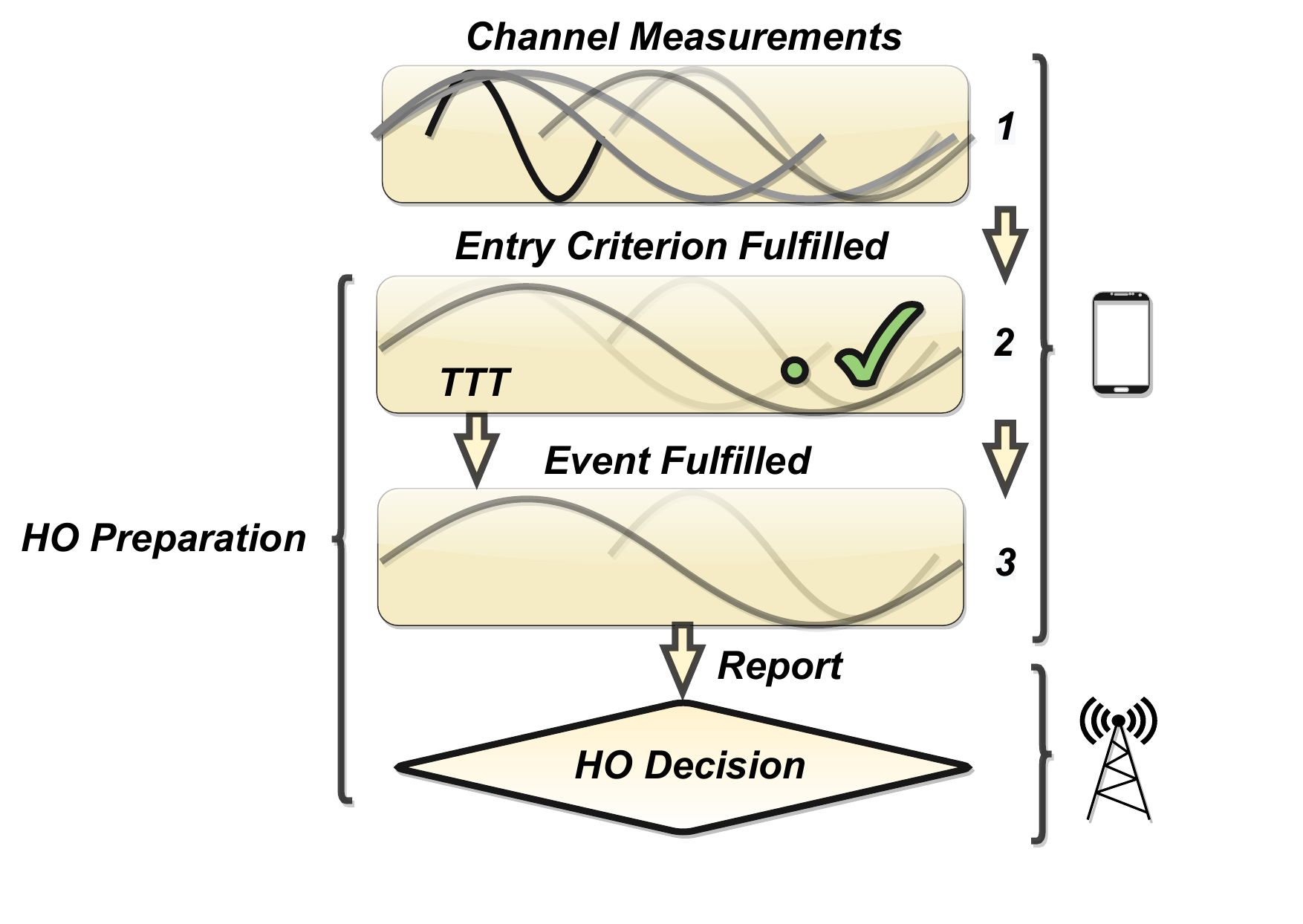}%
		\caption{Interplay between UE and NW during the handover HO preparation phase.}
		\label{fig:HO_procedure} 
	\end{figure}

    \IEEEPARstart{T}{o} ensure seamless user mobility between neighboring cells, the handover (HO) mechanism is defined in the 3GPP specification 38.300 \cite{38300}, from the First Generation (1G) onward. Reliable communication during the mobility of user equipment (UE) is crucial, and HO management is a key capability \cite{Parkvall}. During HO, control messages are exchanged between the UE and the serving Base Station (BS) under predefined conditions. However, since these messages are sent over the air interface, they may be initiated when the radio link faces severe attenuation and various propagation issues such as noise and interference. A robust HO mechanism is essential to maintain user mobility under these conditions; otherwise, user mobility is compromised.
    \newline 
    \indent To address these challenges, each generation of cellular networks has refined the HO procedure while maintaining its core functionality, which consists of three phases: \emph{preparation}, \emph{execution}, and \emph{completion}. The preparation phase, as the initial step of the HO procedure, typically occurs when the signal quality of the serving cell is low and interference from neighboring cells is high. This makes the UE exposed to Handover Failure (HOF) and Radio Link Failure (RLF), therefore, among the three phases, HO preparation is the most vulnerable \cite{3GPP_RAN_WG2}.\newline
	\indent The existing event-driven 5G HO procedure requires the participation of both UE and BS during its preparation phase. In the initial part of this phase, the UE is primarily responsible for measuring the quality of the channel of the serving and neighboring cells and reporting when a measurement event is fulfilled. More precisely, an offset value and a hysteresis value, jointly called the HO margin (HOM), determine when an \emph{entry criterion} of a measurement event is fulfilled, depicted as Step 2 in Fig. \ref{fig:HO_procedure}, where the intrinsic delay of the Time-to-Trigger (TTT) timer bridges Steps 2 and 3. The HOM is the most significant parameter to control the HO decision \cite{38133}. The TTT timer and HOM comprise a tightly coupled setting named HO Control Parameters (HCP) which determine when an HO \emph{event} (HE) is \emph{fulfilled}, depicted as step 3 in Fig. \ref{fig:HO_procedure}, and thus impact the initial timing of an HO preparation phase. For optimal initiation of the HO preparation phase, it is essential to adjust the HO timing to each user's specific mobility pattern and current radio conditions. Fig. \ref{fig:HO_procedure} also illustrates how the traditional HO preparation mechanism assigns a passive and disadvantageous role to the BS, making it unaware of imminent HO events and thus prone to initiate an HO too late.
 
     \subsection{A3 Handover Event}
	 \indent In this section, we examine the core components of an HO event-triggered mechanism, as specified in 3GPP 36.133 \cite{38133}, and clarify how HCPs interact. 
	\vspace{-10pt}
	\subsubsection{Handover Control Parameters}
\begin{figure}[t]
		\centerline{\includegraphics[width=1.03\linewidth, keepaspectratio]{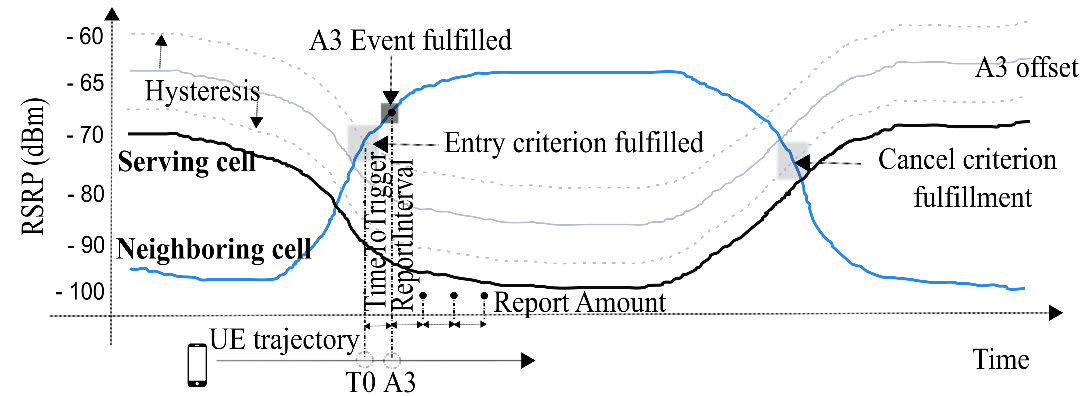}}   	\caption{A3 Event. The quality of neighboring cells exceeds the quality of the serving cell by an offset value. A3 event entry criterion fulfillment (T0) throughout the TTT duration (A3).}
		\label{fig:A3Event}
	\end{figure}

\begin{figure}[t]
		\centerline{\includegraphics[width=1.03\linewidth, keepaspectratio]{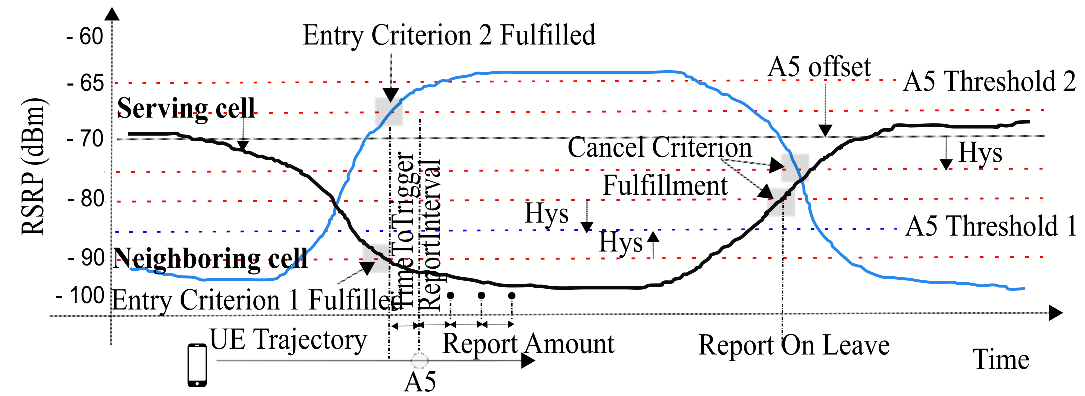}}   	\caption{A5 Event. Only when both entry criteria are satisfied, the UE reports event A5 to gNB.}
		\label{fig:A5Event}
	\end{figure}

	The A3 and A5 events, illustrated in Fig. \ref{fig:A3Event} and \ref{fig:A5Event}, embody the signal quality of the serving cell and neighboring cells using the reference signal received power (RSRP) metric. Event A5 provides a handover triggering mechanism based on \emph{absolute} measurement results. Only the A3 event evaluates a \emph{relative} comparison between the signal quality of the serving cell and that of neighboring cells, making it adaptable to varying network conditions. As we focus on an intra-frequency HO scenario, we chose the most widely used A3 event whose entry criterion fulfillment, hereafter referred to as \textbf{T0}, is given by the inequality (\ref{eqn:RSRP_formula}) where $RSRP_{Target}$ and $RSRP_{Serving}$ are long-term averaged Layer3-filtered measurements from the serving and neighboring cells, respectively. 
 
    \vspace{-18pt}
	\begin{equation}
		\label{eqn:RSRP_formula}
		RSRP_{Target} > RSRP_{Serving}+HOM.
	\end{equation}
	\vspace{-35pt}
    \subsubsection{Handover Margin}
    Any inappropriate HOM settings between low and high may lead to a ping-pong effect or high Radio Link Failure rate. The HOM setting is set at the cell level, which means that all users within the cell will apply the same HOM. Preferably, the adjustment of the HOM settings shall be adapted individually concerning each user's context such as velocity, mobility pattern etc \cite{IndividHCP}. Even though the HOM determines the \textbf{T0} fulfillment it can not be entirely decoupled from the TTT functionality in the context of event-triggered HO optimizations.   
	\vspace{-20pt}
	\subsubsection{Time-To-Trigger}
	Upon \textbf{T0} fulfillment, the UE awaits TTT expiry before reporting HE fulfillment to the BS, hereafter referred to as \textbf{A3}, and as illustrated in Fig. \ref{fig:A3Event}. The TTT timer has been introduced in previous generations of cellular networks and inherently introduces a time delay. If the TTT value is too large, it may cause connection interruption and an HOF. Conversely, too small a value can prevent long delays but lead to an increased HO ping-pong or unnecessary HO. \newline \indent Given the discussed background, sub-optimal HCP settings can negatively impact the optimal timing for HE and reduce the overall HO success rate.
    \subsection{Related Work} 
    The HCP parameters heavily influence the timing of the HO preparation phase, and numerous techniques have been developed to ensure that the HO is initiated at the most optimal moment. The number of potential HO regions inevitably increases in dynamic mmWave environments characterized by reduced cell coverage and multi-beam architecture requiring smaller cell sizes. An HO region is the distance between the HO event trigger point and the Physical Downlink Control Channel (PDCCH) outage point \cite{HO_mgmnt_tutorial}. The handover failure (HOF) rate is directly proportional to the UE mobility speed and inversely proportional to the size of the HO region. The HOF rate can be reduced by expanding the hypothetical HO region through careful tuning of HCP parameters, which must account for varying network deployments, cell sizes, user velocities, and mobility patterns. \newline
    \hspace*{\parindent} Unlike previous research, we dissect the event-triggered mechanism and explain how to distinguish it into two chronological occurrences, advancing the timing of the HO preparation phase. Our machine learning (ML)-assisted method decouples these events by predicting the earliest \textbf{T0} based on changes in the signal patterns of the UE beam measurements.
    From a network perspective, our solution claims insights into steps 2-3 illustrated in Fig. \ref{fig:HO_procedure}.\\
    We briefly shed light on the strengths and limitations of two key optimization techniques that represent the most relevant related research, namely Conditional Handover and Mobility Robustness Optimization.\\\\
    \textbf{Conditional Handover} (Conditional HO), introduced by 3GPP in 5G Release 16 \cite{38300} decouples the base station (BS) preparation and HO execution phases, reducing the number of HOFs by allowing the UE to decide when to initiate the HO. Unlike baseline 5G HO schemes, Conditional HO employs early HO preparation to mitigate the risk of a critical signal quality drop between the UE and the BS. The authors of \cite{ECHO} propose an improved conditional HO scheme that uses trajectory prediction to prepare the BSs along the path of the UE. In contrast, \cite{Conditional_HO_Pred} explores ways to improve early preparation success by predicting the next BS during Conditional HO. These techniques aim to optimize the timing of the HO preparation phase by shifting the responsibility entirely to the UE. However, Conditional HO introduces significant signaling overhead during the HO preparation phase, particularly in dense cell deployments with high HO frequency \cite{Fast_CHO}. However, the Conditional HO technique has some disadvantages and challenges that must be addressed.\\ 
    \emph{Signalling Overhead:} Conditional HO requires the network to pre-configure multiple target cells for a potential handover, which adds complexity to network management. \\
    \emph{HO Decision-Making:} The decision logic for triggering a handover becomes more complex, as the UE has to monitor multiple candidate cells and decide which one is optimal under changing conditions.\\\\   
    \textbf{Mobility Robustness Optimization} (MRO) approaches fall under the HO self-optimization technique family, which aims to automate HCP settings with minimal human intervention. Approaches include optimizing HCP parameters individually, considering trade-offs, or treating them as a unified entity \cite{DataDrivenPredMitigat} - \cite{EricssonRANwithAI}. Studies like \cite{AutoTuningSelfOpt, DynamicHCPAlg} emphasize the need to adapt HCPs in millimeter-wave (mmWave) deployments with dense small cells. These studies propose algorithms to adjust HCPs based on RSRP and UE velocity, continuously refining these parameters after each measurement report. However, despite improvements in performance metrics, these solutions present notable challenges. \\
    \emph{Signalling Overhead:} MRO requires ongoing adjustments to HCPs based on real-time network conditions, which can occasionally result in HO failures or unnecessary HOs. Additionally, MRO solutions often rely on generalized approaches that may overlook the specific UE context such as mobility patterns or velocity, leading to suboptimal performance in certain scenarios. \\
    \emph{HO Decision-Making:} The self-optimization process could lead to either too aggressive or too conservative HO decisions, further contributing to handover failures or an increase in unnecessary handovers.\\
    \emph{Inaccurate Handover Predictions:} MRO algorithms rely on predictive models to optimize handovers. If user mobility patterns or network conditions change suddenly or unpredictably, the system might make inaccurate predictions. \newline \hspace*{\parindent} It is evident that optimizing the timing of the HO preparation phase is a recurring focus in much of the research conducted in this area. In the following sections, we explain how the proposed \emph{Early-Scheduled Handover Preparation} (ESHOP) scheme addresses the advantages above and limitations identified in related research, as well as those within the ESHOP framework itself. Regarding MRO techniques, continuous adjustment of HCPs requires frequent signaling in the downlink via measurement radio resource reconfiguration, which significantly increases power consumption on the UE side \cite{38331}. This issue becomes particularly pronounced at high UE velocities and in small cell deployments with frequent HOs. Furthermore, these solutions assume that the UE's velocity is known, a parameter that is typically not tracked by cellular networks. Our solution also relies on dedicated signaling toward the UE and is sensitive to sudden and unpredictable changes in UE mobility patterns. Unlike traditional approaches, however, our solution can learn from measurement data, improving handover robustness even in the face of unexpected events. \\ When optimizing the HO preparation phase, Conditional HO complicates decision-making by triggering multiple target cells. In contrast, our study employs a different approach to optimize the timing of the HO preparation phase and reduce signaling overhead. Instead of explicitly estimating individual UE paths or velocities using conventional wireless channel modeling, we utilize a technique that associates a series of radio channel measurements with physical locations through channel fingerprinting. These fingerprinted features, based on each user’s trajectory and velocity, enable us to analyze the time series of relationships between these variables, forming the foundation of our study. This approach allows the ESHOP scheme to trigger HO preparation in a just-in-time manner.
    \subsection{Contributions}
    \begin{itemize}
        \item \emph{Predictive Timing}: By accurately predicting the timing of the \textbf{T0} fulfillment, the ESHOP scheme allows the network to initiate HO preparation in advance and ensures that the preparation phase begins at the most appropriate time. This proactive approach contrasts with traditional reactive methods that wait for the subsequent \textbf{A3} fulfillment to be met and reported by UE before initiating HO preparation.\vspace{5pt}
        \item \emph{Enhanced HO Regions}: The proposed ESHOP scheme proactively expands the hypothetical HO region by initiating the preparation phase earlier. This expansion allows for more time to manage the HO process, thereby minimizing the risk of users experiencing signal degradation or loss of connectivity during the HO. This is particularly beneficial in dynamic mmWave environments characterized by small-cell deployments.\vspace{5pt}
        \item \emph{Dynamic HO Preparation}: The ESHOP scheme dynamically adjusts the timing of the HO preparation phase. This user-centric approach enables flexibility in accommodating rapid changes in the radio environment, thereby enhancing the robustness of the HO process.
    \end{itemize}
    \section{SIMULATED MODEL SETUP}
    \label{System} 
\begin{figure}[!b]
            \centerline{\includegraphics[width=\columnwidth, keepaspectratio]{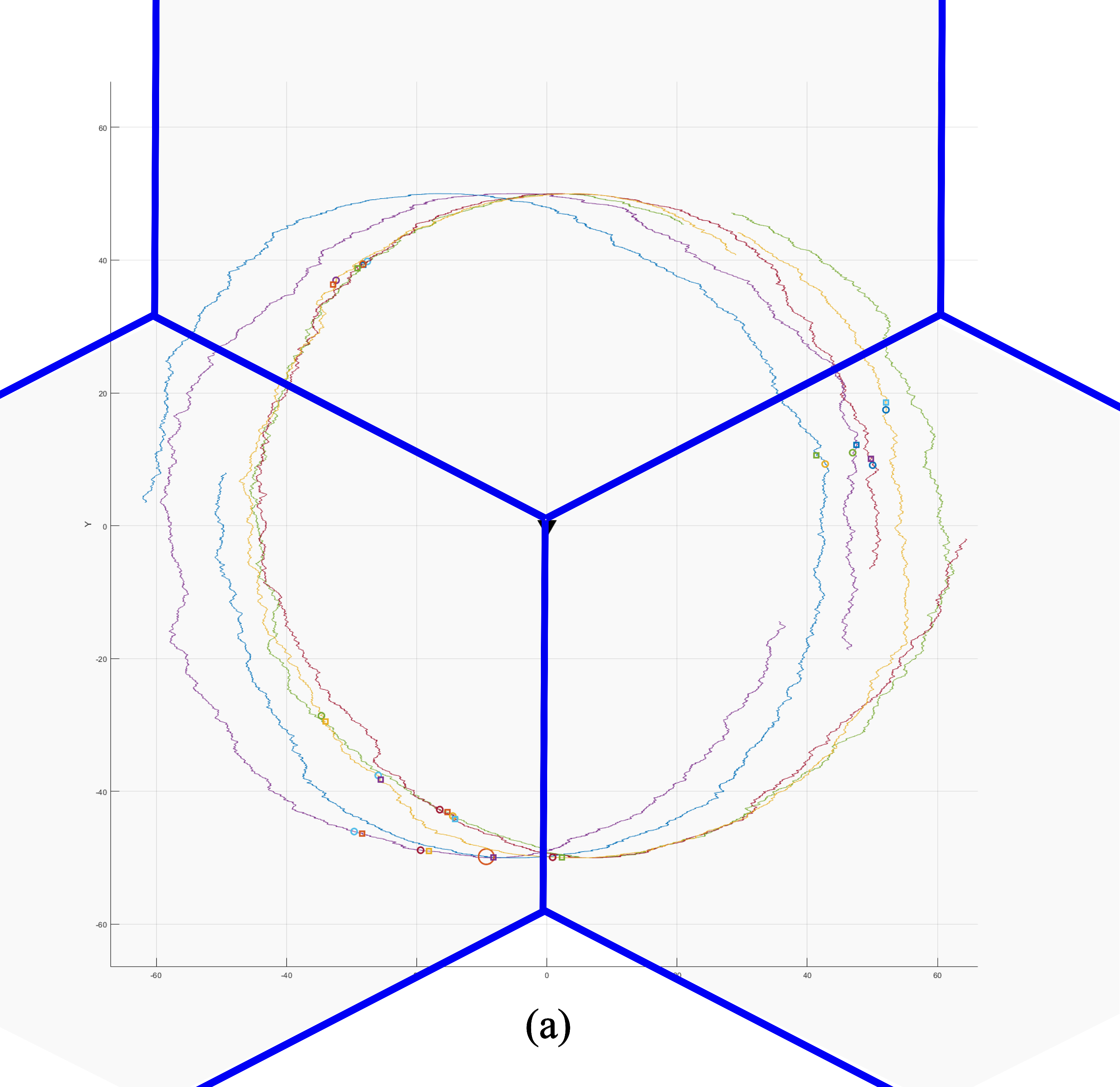}}
        \centerline{\includegraphics[width=\columnwidth, keepaspectratio]{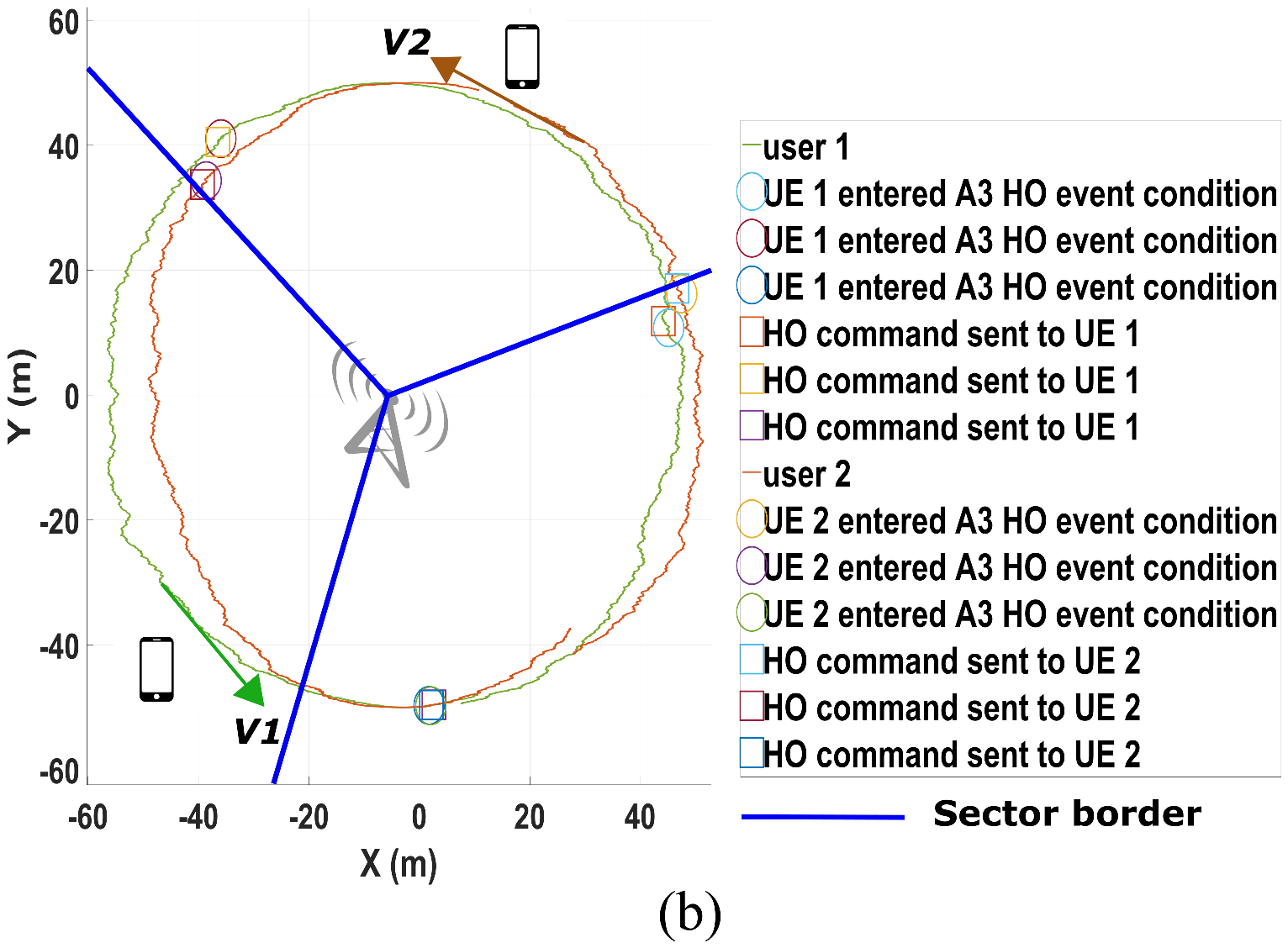}}
        \caption{Top view of random individual trajectories generated by five distinct users (a), 5G NR site deployment with the base station located at the corners of three adjacent cell sites, each shaped as a hexagon and depicted in dark blue (b). Users move along individually randomized circular trajectories at constant velocities of 25 m/s (v1) or 31 m/s (v2).}
        \label{fig:Sitedeployment}
    \end{figure}

    To demonstrate the primary goal of this investigation, which is the feasibility of using beam measurements for HO predictions, this study employs an extremely simplified mobility model as a proof of concept. We acknowledge that the simulated system does not represent a typical urban 5G network deployment, such as a crowded metropolitan area. Despite its simplicity, the system model effectively presents fundamental HO-related issues in a small-cell network deployment, and the simulations provide a realistic HO procedure using a radio system that supports detailed beam management that is compliant with standardized 5G NR systems at mmWave frequencies \cite{38104}. We simulate a commercial-grade Phased Array Antenna Module in a BS operating at a center frequency of 28 GHz over a 100 MHz bandwidth. The 5G NR frame numerology is set to a 120 kHz subcarrier spacing with a slot duration of 125 $\mu$s (8000 slots per second). We simulate a three-dimensional area with a single site containing three cells, with the BS deployed in the center of three cells shaped as hexagons, as shown in Fig. \ref{fig:Sitedeployment} (b). 
    Due to the small cell deployment, the coverage extends only over a fraction of each hexagon, since dense cell deployments typically allow limited freedom of movement within a cell. As depicted in Fig. \ref{fig:Sitedeployment} (a), the UE trajectory follows a circular but individually randomized path centered around the BS with a 50-meter radius, modeling a dense small-cell deployment that results in rapid and frequent HOs. The circular trajectory simultaneously presents a variety of instantaneous movement angles in relation to the base station position, emulating a mix of user movement patterns in a practical deployment. UEs appear along the circular trajectory at random starting points, causing variability in the timing of the next HO.

\begin{figure}[t]
		\centerline{\includegraphics[width=1.1\columnwidth, keepaspectratio]{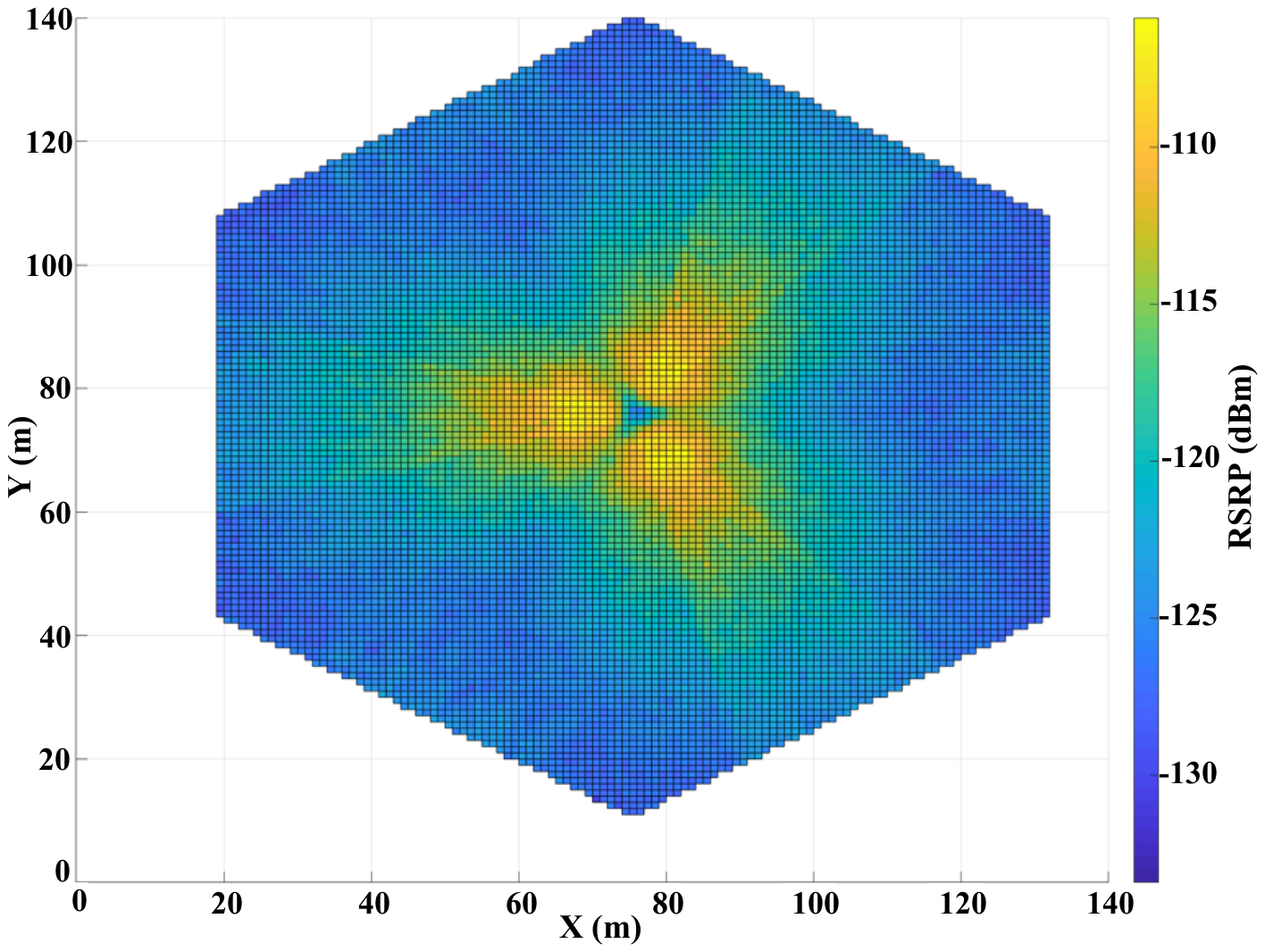}}
		\caption{2D projection of RSRP distribution generated by the 4x3 SSB wide beam pattern.}
		\label{fig:4x3_SSB_Envelope}
	\end{figure}

    \newcolumntype{P}[1]{>{\centering\arraybackslash}p{#1}}
	\begin{table}[b]
		\caption{Simulated handover parameters}
		\centering
		\renewcommand{\arraystretch}{0.2}
		\resizebox{\linewidth}{!}{%
			\begin{tabular}{|P{.25\textwidth}|P{.25\textwidth}|}
				\topline
				\headcol Parameter & Value \\
				\midline
				measurementType & A3  \\
				\midline
				timeToTrigger & 0.04 (s)  \\
				\midline
				hysteresis & 0 and 1 (dBm)  \\
				\midline
				reportInterval & 0 (s) \\
				\midline
				reportAmount & 1 \\
				\midline
				offset  & 3 (dBm) \\
				\midline
				measurementQuantity  & RSRP \\
				\bottomlinec
			\end{tabular}
		}
		\label{table:simparams}
    \end{table}
    To reveal the necessity of initiating the HO preparation at its earliest stage, we simulate users traveling through multiple cells, spending only a short time within each cell along its trajectory. The circular trajectory ensures that the user crosses all three cell borders during a single simulation duration, with each crossing occurring at slightly different angles and times, depending on the randomness of the UE's movement along the path. Cell border crossings are influenced by cell coverage overlap and potential HO opportunities. Users move at constant speeds of either 25 or 31 m/s, which significantly alters the dynamics between the BS and the user in terms of rapidly changing channel quality which is captured by the measurement reports sampled periodically at constant time intervals of 40 ms. During the simulations, the UEs are in radio resource control connected mode and engage in active UL-centric signaling towards the BS. The lifetime of each UE is recorded from the start to the end of the simulation. Parameter settings were evaluated for different UE loads using the HO-related parameters listed in Table \ref{table:simparams}. Multiple seeds were used to ensure statistical confidence. Both Line-Of-Sight (LOS) and Non LOS propagation scenarios were included, as per the guidelines in \cite{38901}. As this study focuses on time predictability using historical RSRP values at standardized mmWave frequencies, bands outside these specified frequencies are beyond the scope of this study, and the conclusions should not be extrapolated to other frequencies. \newline
    \hspace*{\parindent} From a UE perspective, an NR cell is defined by the physical transmission of a specific Synchronization Signal Block (SSB), which contains a physical cell identifier at a particular frequency. The BS transmits SSB beams periodically, with a $T_{SS}$ of 10 ms. These beams are arranged in a grid of 3 in azimuth and 4 in elevation, resulting in $N_{SSB} = 12$ beams, as depicted in Fig. \ref{fig:4x3_SSB_Envelope}. 
    The SSB beams are static and wide, always pointing in the same direction, forming a grid that covers the entire cell area. This makes them suitable for cell-level mobility evaluations. An HO occasion is dynamically defined by the rapidly changing mmWave radio environment conditions, fingerprinted in a time series of SSB beam measurements specific to each UE's movement pattern on UE trajectory, location and velocity. This is particularly notable as individual circular trajectories may be separated by tens of meters, as illustrated in Fig. \ref{fig:Sitedeployment} (a). Demonstrating robust behavior of the method at higher speeds also suggests corresponding or further improved performance at lower speeds, assuming that model is trained with corresponding data patterns at various speeds. As UEs move around the BS, they search for and measure the qualities of these beams, maintaining a set of candidate beams from multiple cells. A combination of a physical cell identifier (cell ID) and beam identifier (beam ID) differentiates beams from each other.
 
    \section{PROPOSED ESHOP SCHEME}
    \label{ESHOP}
\begin{figure}[t!]
    \centering 
    \includegraphics[width=1.0\linewidth]{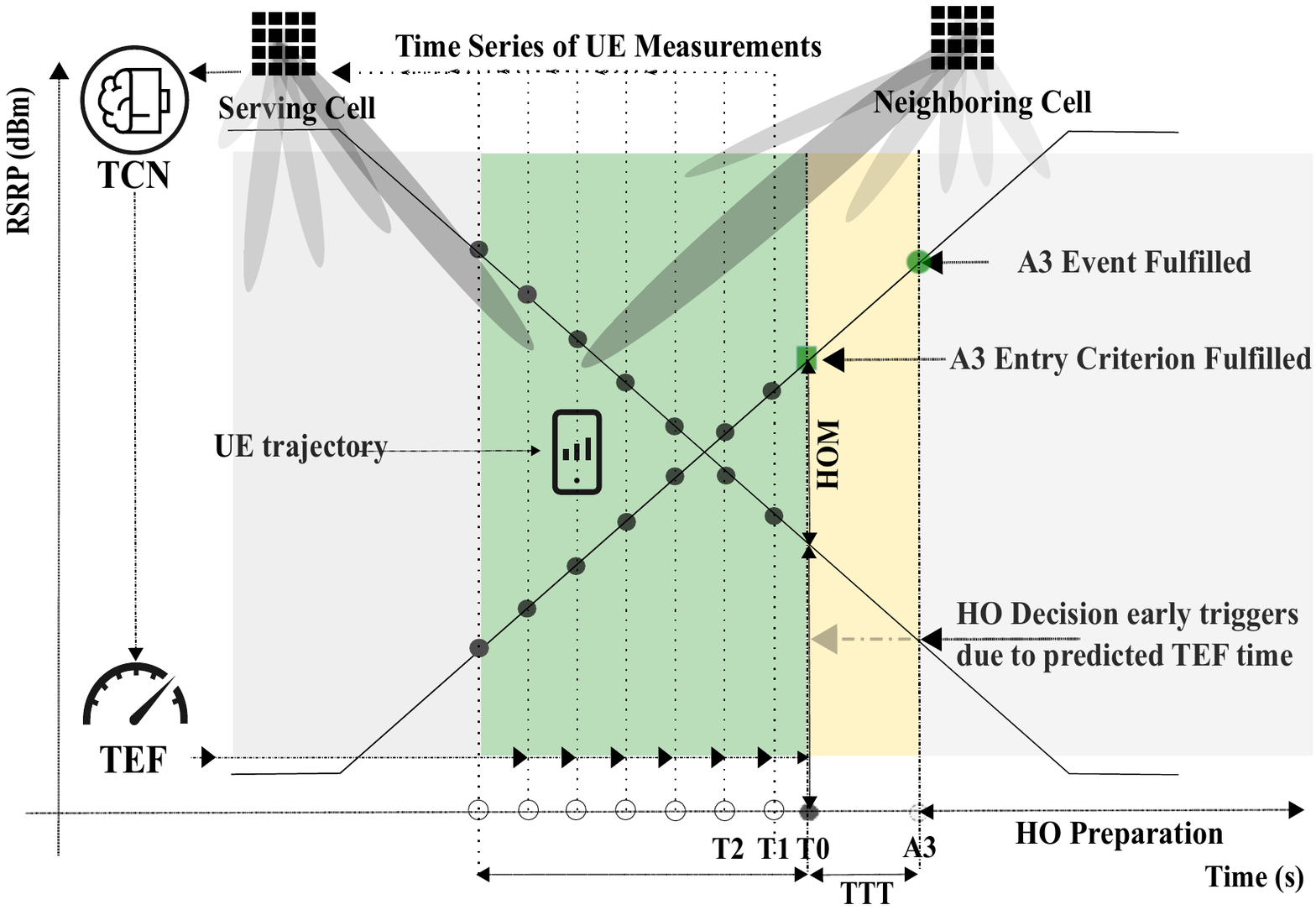}
    \caption{Principals of ESHOP scheme and its most significant component, Time To Entry Criterion fulfillment (TEF).}
    \label{fig:ESHOP_scheme} 
    \end{figure}

    In this section, we examine the incentives underlying the introduction of ESHOP and explain the application area of our TCN-driven model, which offers several benefits worthy of close evaluation. Before the UE triggers an HE, it measures the signals of the serving cell and neighbor cells over the 5G NR air interface evaluating whether any measured signals satisfy the \emph{entry criterion fulfillment} of a HE, \textbf{T0}, as highlighted in green in Fig.  \ref{fig:ESHOP_scheme}. The ESHOP aims to increase the robustness of the HO preparation phase by utilizing the time window between \textbf{T0} and \textbf{A3}, where \textbf{A3} marks the start of both the potential HO region and the legacy HO preparation phase. These two events are time-bridged by the TTT value, as highlighted in yellow in Fig. \ref{fig:ESHOP_scheme}. 
    The most significant component of the ESHOP scheme is the Time to Entry Criterion Fulfillment (\textbf{TEF}), i.e. the remaining time until \textbf{T0}. Being a predictive measure, \textbf{TEF} estimates the time at which the \textbf{T0} will be met allowing the network to initiate HO preparation well in advance, thereby reducing the likelihood of HOFs and improving overall network performance. The ESHOP scheme enables a user context-aware detection of incoming \textbf{T0} occasions. It establishes a new starting point for both the potential HO region and the HO preparation phase, scheduling them earlier by exactly the TTT duration.
    As developed in detail later in Section \ref{Discussion}, we aim to contain the HO decision procedure between source and target cells within the TTT window thereby early-triggering the HO preparation phase. The ESHOP relies on the capability of the data set to capture the fingerprinted features based on each user's trajectory and velocity and allows us to study the time series of relationships between these. 
    As shown in Fig.  \ref{fig:A3Event} and \ref{fig:ESHOP_scheme}, the HO hysteresis setting, as part of the HOM, impacts the timing of \textbf{T0} and pushes it along the time axis and the UE trajectory, e.g, when set negative it triggers HO preparation earlier risking a ping-pong effect. Conversely, a late-triggered HO preparation risks RLF and HOF. An obvious tradeoff between HO failures and ping-pongs is strongly related to the UE velocity whereas an expansion of the potential HO region is a possible solution  \cite{5G_URLLC}. Note that in the leading boundary of a traditional HO region, the \textbf{A3}, may be pushed in both directions, along the UE's trajectory depending on the hysteresis setting. The same is valid for the new leading boundary \textbf{T0} separated from \textbf{A3} via the TTT timer. The latter alludes to the essence of the ESHOP scheme and the underlying data set containing fingerprinted information about e.g.; hysteresis setting as well its co-relation with the UE's trajectory and velocity. This fact alleviates the need for additional optimizations of HOM parameters and allows for focusing on utilizing the TTT timer's duration for signalling time savings. Purposely, TTT duration was kept constant throughout this study. As a gNodeB remains unaware of an imminent \textbf{A3} occasion in the legacy HO procedure, the ESHOP scheme addresses this by assigning an \emph{active} role to the gNodeB, enabling it to predict the remaining time to \textbf{TO} via the ML-inferred \textbf{TEF} metric.

    \section{ENHANCEMENTS FOR HO ROBUSTNESS}
    \label{Discussion} 
    	\begin{figure}[b]
        \includegraphics[width=1.0\linewidth]{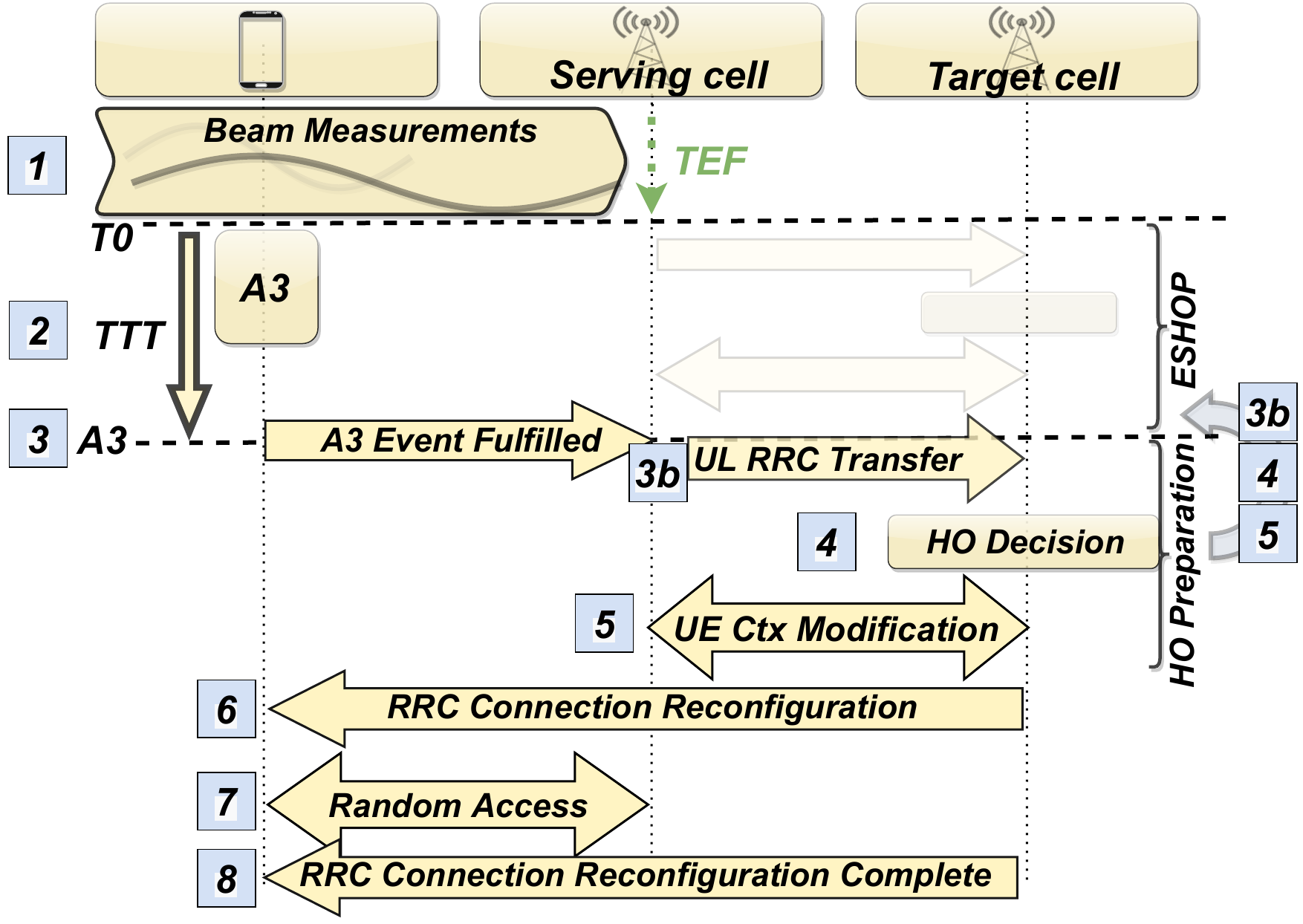}
		\caption{Legacy intra gNodeB handover procedure and the proposed ESHOP scheme.}
		\label{fig:HO_Intra_flowscheme}
	\end{figure}

  	\begin{figure}[b]
		\includegraphics[width=1.0\linewidth]{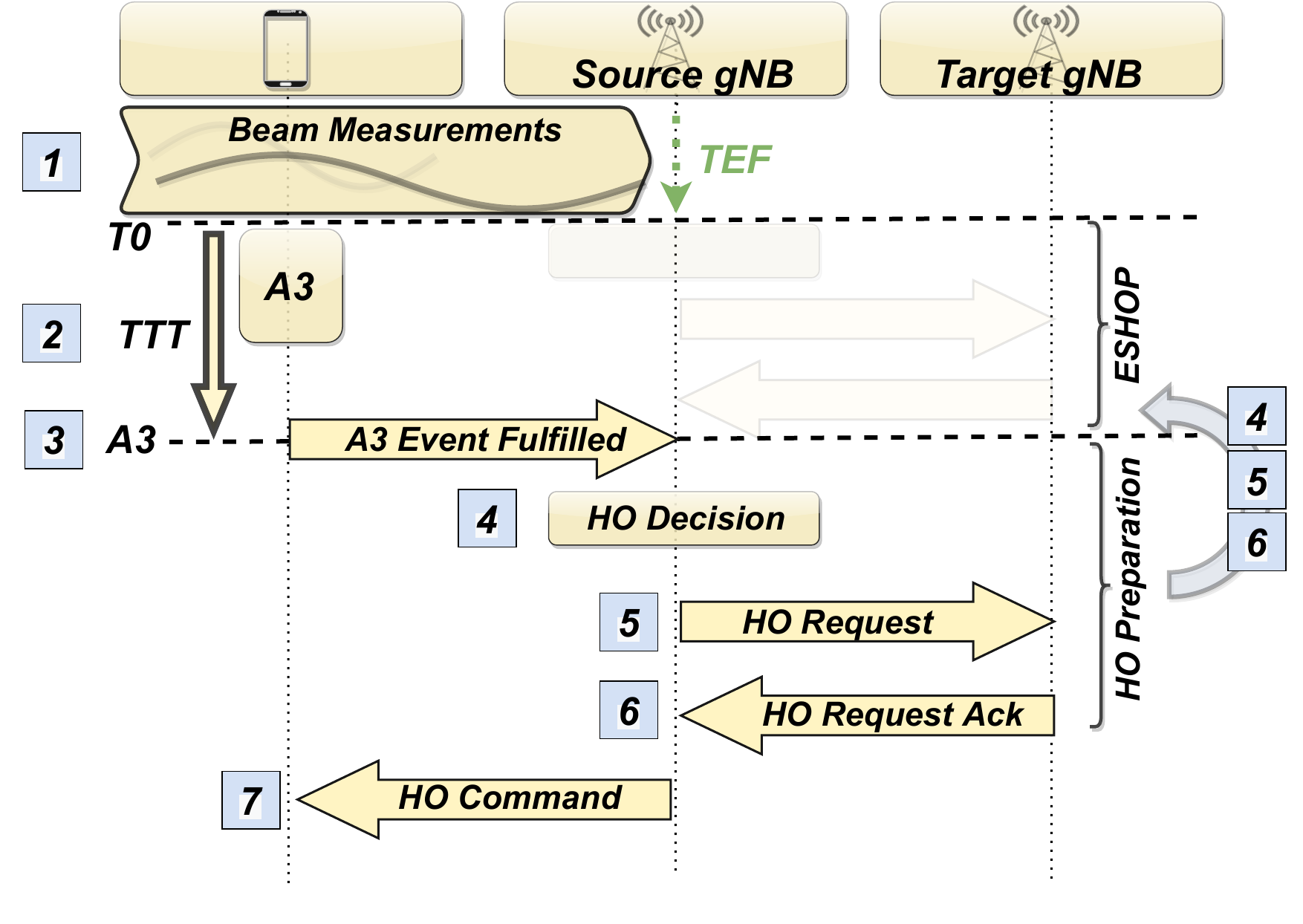}
		\caption{Legacy inter gNodeB handover procedure and the proposed ESHOP scheme.}
		\label{fig:HO_Inter_flowscheme}
	\end{figure}
    This section demonstrates how the ESHOP scheme can be applied on the network side to enhance the robustness of the 5G HO. The HO procedure involves various levels of internal signaling between network nodes, with intra-gNB communication being the simplest in terms of signaling complexity. Figures \ref{fig:HO_Intra_flowscheme} and \ref{fig:HO_Inter_flowscheme} illustrate the general signaling order for both intra-gNB and inter-gNB communication. Upon fulfillment of the A3 event, the HO preparation is triggered. As part of the HO decision, the serving cell exchanges the UE's context with the candidate cells, and based on available resources, the target cell performs admission control to determine if it can accommodate the incoming UE. If admission is granted, an HO acknowledgment is sent to the source node, including an HO command conveyed in the downlink to the UE. As described in Section \ref{ESHOP}, the ML model continuously predicts the remaining time approaching \textbf{T0} via \textbf{TEF} metric, represented as a countdown. The different stages of the proposed ESHOP scheme for intra-gNB and inter-gNB HO scenarios are illustrated in Figures \ref{fig:HO_Intra_flowscheme} and \ref{fig:HO_Inter_flowscheme}. In the following, we exemplify the ESHOP scheme depicted in Figure \ref{fig:HO_Intra_flowscheme}; the same principles apply to Figure \ref{fig:HO_Inter_flowscheme}.
 
	\begin{itemize}
		\item \textbf{Step 1}: The BS continuously receives beam measurement reports for the serving cell and neighbor cells for each active UE. With the aid of the incoming measurement reports, the ML model predicts the remaining \textbf{TEF} time based on its training experience and evaluates whether to proceed with an early triggering of the HO preparation phase. The incoming HO event criterion fulfillment ideally results in a continuously decreasing \textbf{TEF} value, approaching \textbf{T0}. Once \textbf{T0} is reached, the next step is initiated. 
		\item \textbf{Step 2}: An \textbf{A3} event is expected to be reported by the UE after TTT expiry. The BS utilizes the TTT duration for early activation of the HO preparation phase, starting with the HO decision-related procedures such as steps 3b, 4, and 5. At this stage the source gNB decides whether to prepare the target cell in advance. If the target cell is prepared, the next step is initiated.
		\item \textbf{Step 3}: Upon TTT expiry, an A3 event is expected to be reported by the UE. \newline \textbf{Note:} In case the ESHOP prediction fails suggesting that UE does not report an A3 event fulfillment, the subsequent steps are not executed, and the current ESHOP scheme is aborted, allowing the BS to fall back to the legacy HO preparation procedure.
		\item \textbf{Step 6-8}: Based on already prepared HO decision-related signalling in the previous step, the remaining signalling of the HO procedure is carried out.
	\end{itemize}
	The ESHOP demonstrates the capability to mitigate HO failures induced by often sub-optimal HOM settings. Considerable time savings are achieved during the HO preparation phase by performing the most time-consuming signalling during the TTT time window. The remainder of this paper presents the feasible time savings achieved through this approach, paving the way for a higher success rate in the HO preparation phase, especially in the mmWave frequency bands \cite{3gpp_study}. For the ESHOP scheme in Fig. \ref{fig:HO_Intra_flowscheme}, we note that including the signaling step 6 within the scope of the ESHOP scheme would cause the TTT abortion according to \cite{38331}. The same applies to Step 7 in Fig. \ref{fig:HO_Inter_flowscheme}. Therefore, we confine the signalling optimizations within the proposed ESHOP scheme signalling scopes. Nevertheless, the BS retains the flexibility to cancel the ESHOP scheme at any time and revert to the legacy HO preparation procedure if necessary. In addition, when ESHOP fails to predict the fulfillment of the A3 event (Step 3), even though the BS can fall back to the legacy procedure, this results in unnecessary HO preparation signaling which is the most significant disadvantage of the proposed HO scheme. \newline
    \hspace*{\parindent} Finally, the findings of this study are limited to the simulated dataset generated based on the system described in Section \ref{System}. In contrast, real network deployment would necessitate a larger dataset encompassing various mobility patterns adding more complexity when training the proposed ML model. However, the small-cell network deployments allow only for limited mobility patterns, confined to a specific coverage area, lending credibility to our model setup despite its simplicity.
    
    \section{MACHINE LEARNING FRAMEWORK} \label{MLModels}
    This section clarifies how ML algorithms can beneficially discover cell relationships between serving and neighboring cells to improve the robustness of the 5G HO preparation phase. Robust mobility management in advanced 5G deployments is challenging due to the unpredictable nature of user mobility patterns. To address this, machine learning (ML) algorithms have proven efficient in analyzing traffic and network data, and they are expected to be essential for improving 5G performance and robustness. ML-based technologies' ability to optimize parameters across multiple layers and identify patterns over complex time series has garnered significant attention in the wireless industry, as they hold the potential to revolutionize wireless network design and deployment. In realistic scenarios involving mobility, either in the propagation channel or due to UE movement, a massive number of signal observations are generated at each port of the gNodeB's MIMO antenna array. Consequently, the radio access network acquires, computes, and processes substantial amounts of data between layers 1 and 3. This implies that ML is ideally applied at higher system layers, utilizing signal observations from layer 1. For real-time HO prediction, it is preferable to select ML algorithms that can handle compute-intensive problems without compromising the baseband processing capacity. Thus, we chose to explore Temporal Convolutional Networks (TCN).
	\subsection{TEMPORAL CONVOLUTIONAL NETWORK} 	
	One significant application of neural networks is sequence modeling, specifically time series analysis, which involves capturing temporal structures in data for the purpose of making time-series predictions. Temporal Convolutional Networks (TCNs) excel in prediction tasks that involve time series data with complex patterns, making them an ideal choice over recurrent neural networks (RNNs). TCNs offer several advantages: they avoid the common drawbacks of RNNs, such as the exploding/vanishing gradient problem and inadequate memory retention. Additionally, TCNs enhance performance by enabling parallel computation of outputs, unlike RNNs. A key feature of TCNs is their causal nature, which ensures that an element in the output sequence relies only on preceding elements in the input sequence. This causality allows for direct conclusions between input and output, something typically not achievable with most machine learning architectures. TCNs are implemented as residual blocks, which further boosts their learning capabilities and enables them to outperform other deep learning networks. \newline
    In the following, we explore the basic building blocks that a TCN consists of, and how they all interplay.
    \subsubsection{Sequence modeling}
    A sequence modeling network is any function given by
	\begin{equation}
	\label{seq_model}        
        \boldsymbol{f} : \mathbf{X}^{n} \xrightarrow{} \mathbf{Y}^{n}
    \end{equation}
    that can be described as a function $\boldsymbol{f}$ that maps a given input sequence $\mathbf{x} \in \mathbb{R}^{n} = [x_0, x_1, ..., x_{n-1}]$ to a corresponding output sequence $\tilde{\mathbf{y}} \in \mathbb{R}^{n} = [y_0, y_1, ..., y_{n-1}]$  such that
    \begin{equation}
	\label{seq_function} 
         \tilde{\mathbf{y}} = \mathbf{f}(\mathbf{x}) 
    \end{equation}
    In predicting task, each element $y_t$ for a specific time index $t$ ($0\leq t \leq n$) can be calculated based on the input vector $\hat{\mathbf{x}}  \in \mathbb{R}^{t} = [x_0, x_1,..., x_{t-1}]$ that collects data that has been previously observed. In other words, the function $\boldsymbol{f}$ is \emph{causal} which means that it does not depend on any future input $x_{t+1}$. In this paper, a neural network aims to solve a sequence modelling task so that the predicted output $\tilde{\mathbf{y}}$ approaches its ground truth $\mathbf{y} \in \mathbb{R}^{n}$. To measure the prediction quality, the following loss function $L_{RMSE}$ is applied 
	\begin{equation}
	\label{seq_function} 
        L_{RMSE}(\mathbf{y}_t, \tilde{\mathbf{y}}_t)) = \sqrt{\sum_{i=1}^t \frac{(y_i - \tilde{y}_i)^2}{t}},
    \end{equation}
    where RMSE is defined as in (\ref{eqn:rmse}).
    \subsubsection{Causal Convolutions} 
    The TCN generates an output of the same length as the input and no data exposure from the future into the past time steps is allowed. The basic convolution relies on a causal input and filter, which makes it inappropriate for sequence modeling tasks, since the convolution operation depends on future time steps. The architecture of TCN is an extended model of a one-dimensional (1D) convolutional neural network (CNN) consisting of stacked convolutional layers and can be described as
    \begin{equation}
	\label{causalConv} 
        F(x_t) = \sum_{p=0}^k f[p]\hspace{1pt}x[t - p],
    \end{equation}
    where $k$ is the size of the filter with an input sequence of length $n$ that returns a sequence of length $n-k+1$ applied at time $t$.
    The zero padding of length $k-1$ appended at the beginning of the sequence ensures the equal length of input and output. In other words, a casual convolution is used to prevent leakage from the past into future steps.
    \vspace{-20pt}
    \subsubsection{Dilated Convolutions}
\begin{figure}[t!]
		\centering  
		\includegraphics[width=0.9\linewidth]{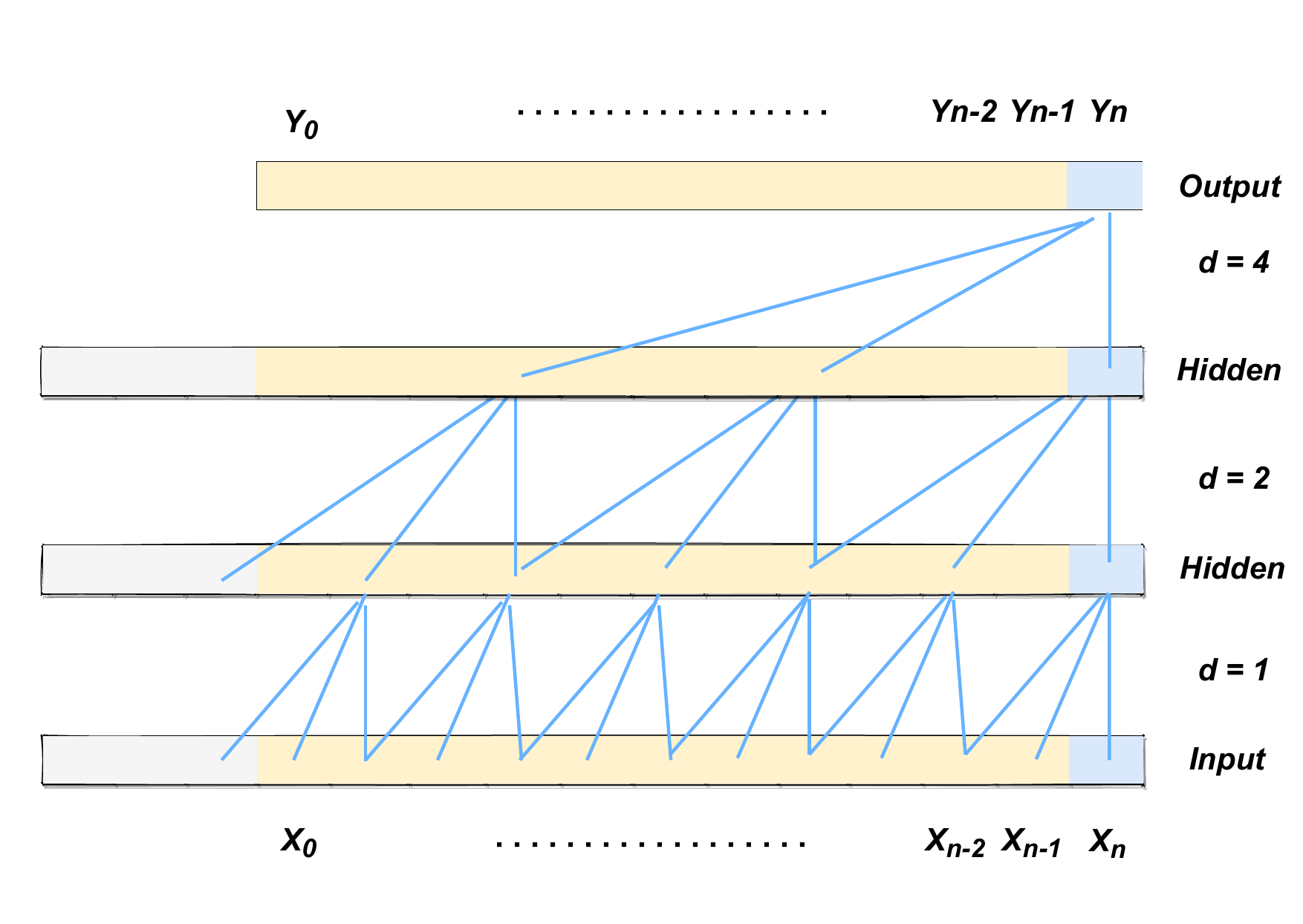}
		\caption{A dilated causal convolution with dilation factors d = 1, 2, 4 and filter size k = 3.}
		\label{fig:TCN} 
	\end{figure}

    A dilated convolution is based on the causal convolution model with a slight modification via the so-called dilation factor $d$ as defined below.
    \begin{equation}
	\label{dilatedConv} 
        F^{\prime}(x_t) = \sum_{p=0}^k f[p]\hspace{1pt}x[t - (d \cdot p)].
    \end{equation}
    A dilated convolution, as shown in Fig. \ref{fig:TCN}, allows a network to understand the dependencies of previous steps and exponentially increase the receptive field by expanding the dilation factor over all layers.
    \vspace{-20pt}
    \subsubsection{Residual Connections} 
\begin{figure}[htbp]
		\centering  
		\includegraphics[width=0.83\linewidth]{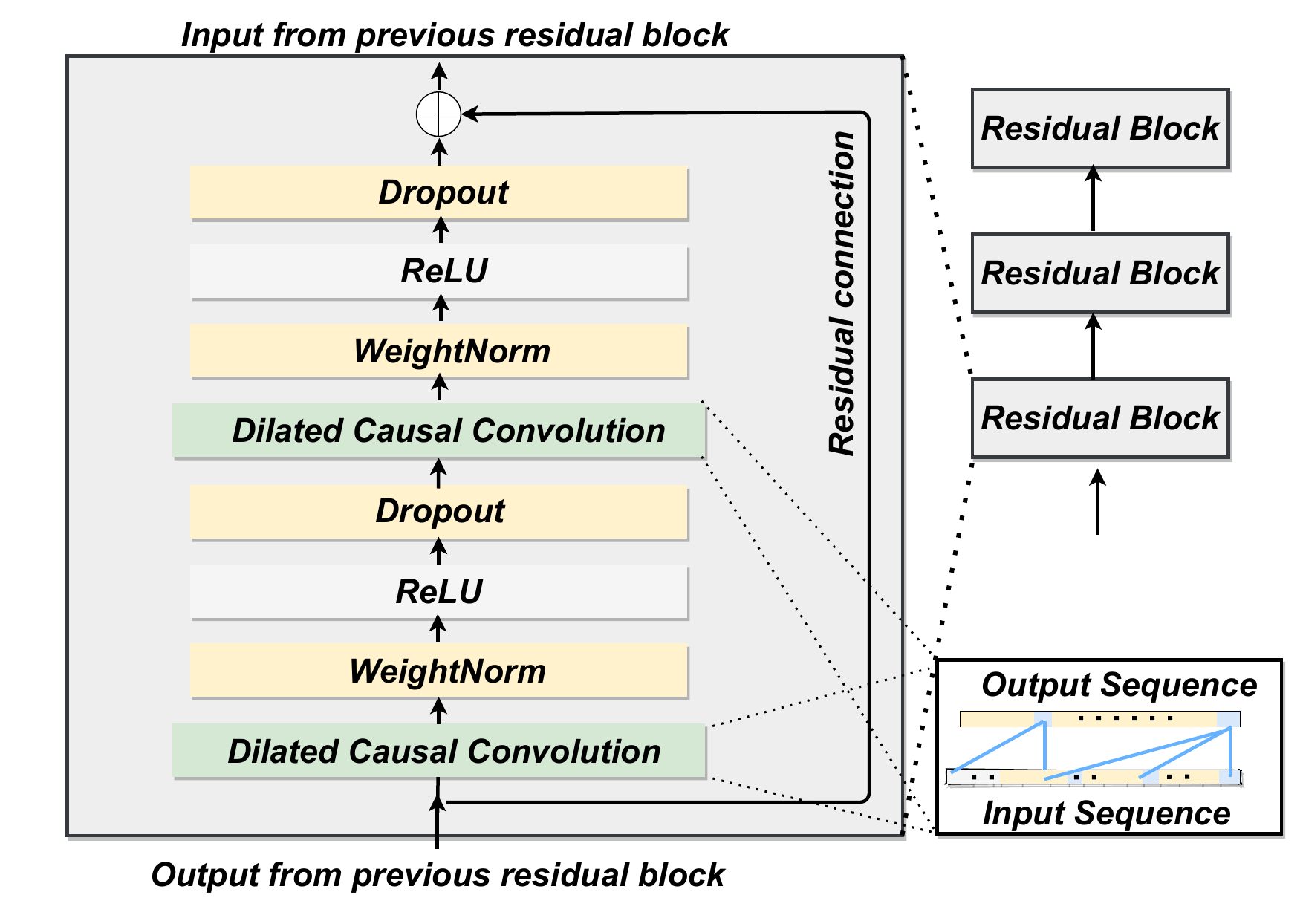}
		\caption{Residual Convolutional Block.}
		\label{fig:ResBlock} 
	\end{figure}

    The mechanism of a residual block is illustrated by Fig.~\ref{fig:ResBlock}. In accordance with Fig.~\ref{fig:ResBlock}, the output of the network $\tilde{\mathbf{y}} \in \mathbb{R}^{n}$ can be expressed as      
    \begin{equation}
	\label{resBlock} 
        \tilde{\mathbf{y}} = f_{relu}(\boldsymbol{x} + \mathbf{F}(\boldsymbol{x})),
            \end{equation}
    By applying  the activation function $f_{relu}$ to the each element $x_i$ in $\mathbf{x}$, we get 
    \begin{equation}
         f_{relu}(x_i) = max(0,x_i)
    \end{equation}
    \vspace{-24pt}
    \section{RESULTS AND EVALUATION}
    We explore the potential of the ESHOP scheme using data generated from extensive simulations conducted within the system framework described in Section \ref{System}. The simulated channel quality estimator for Layer 3 RSRP incorporates filtering based on the SSB beam-specific configuration, as part of the higher-layer radio resource management in 5G NR, following 3GPP standards \cite{38300} \cite{38331}. This filtering aims to eliminate the effects of fast fading and disregard short-term variations. The resulting measurement reports are sampled periodically at predefined time intervals of 40 ms resolution, providing averaged long-term RSRP measurements. 
            \begin{figure}[b!]
		\centering  
		\includegraphics[width=0.98\linewidth]{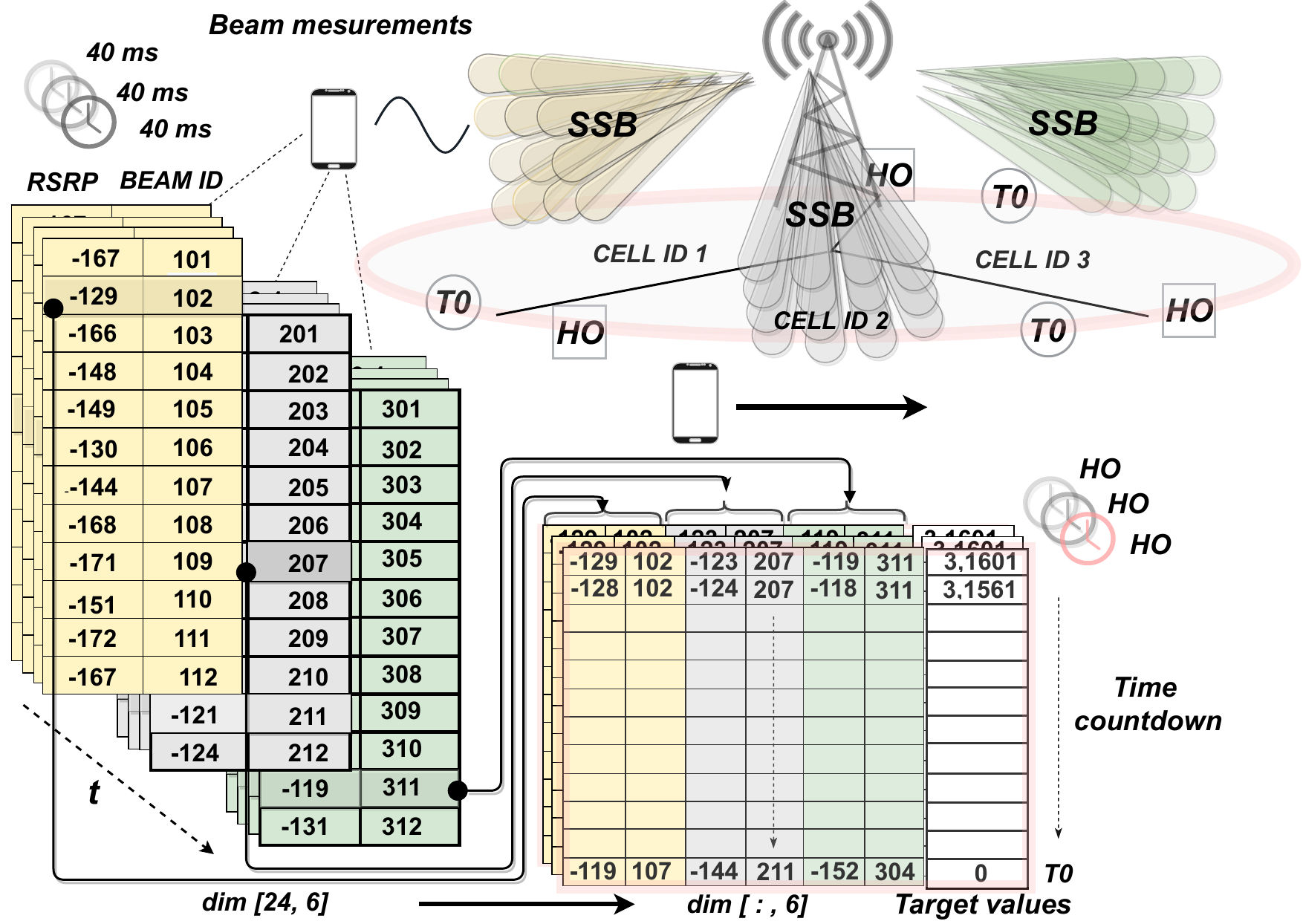}
		\caption{Data set structure. L3-filtered RSRP beam measurements collected with a 40 ms reporting periodicity contain 12 RSRP values for each SSB beam. Between two HO occasions, these 40 ms measurement snapshots are reduced to the strongest beam for each cell, forming a time series of channel measurements including the corresponding time to the next HO occasion as the prediction target. }
		\label{fig:data set} 
	\end{figure}%
    Each beam measurement report comprises 12 pairwise samples of the best RSRP and beam ID per cell. Initially, the time series of measurement reports yields a dimension of 3 cells x 12 beam ID/RSRP pairs, collected from the serving and the two neighboring cells. 
    Subsequently, the dataset is reduced to an input vector containing only 6 features, such as the best RSRP with the corresponding SSB beam ID for each of the three cells, as illustrated in Fig. \ref{fig:data set}. 
    All UE measurements are logged with the UE's simulated lifetime timestamps. Throughout each UE's simulated lifespan, spanning from the simulation's start to its end, we document the occurrence of \textbf{T0}, which is then converted into a countdown format. During ML model training, these timestamps serve as the prediction target, indicating the time remaining until the next \textbf{T0} event, presented as a countdown. Our evaluation is centered on enhancing the robustness of the HO preparation phase. Consequently, we restrict data collection to the moment of HO command transmission, disregarding the actual HO outcome. Additionally, if \textbf{T0} is not maintained for the entire duration of the TTT timer, leading to an A3 event, the associated measurements are excluded. Only measurements preceding the actual start of the HO procedure are considered. 
    It's important to note that our approach, which relies on historical RSRP data, should not be seen as a limitation of our system or the proposed contribution. By incorporating historical RSRP data alongside the corresponding beam ID, our method ensures the effective linkage of the UE's trajectory and velocity with the measured attributes of the received signal. Given that the SSB beams remain static in terms of horizontal and vertical radio coverage, this approach inherently captures fingerprinted HCPs within the dataset, particularly those settings that have successfully satisfied the \textbf{T0} occasion.
 
    \subsection{TCN MODEL PERFORMANCE}
	\label{Resultat}
	\hspace*{\parindent} The employed, relatively simple ML model, consists of a single TCN layer with a kernel size of 11 and incremental dilations with sizes $1, 2, 4, 8, 16, 32, 64$. This TCN layer is followed by three dense neural network layers of size 32, 16 and 8. There are only 590209 total parameters that need to be trained in this network. To evaluate the performance of the ML model the following metrics were used. Residual Mean Square Error R-Squared $R^2$ is calculated by comparing the Sum of Squares of Errors (SSE) to the Total Sum of Squares (SST) (\ref{eqn:r2}), Explained Variance Score (EVS) (\ref{eqn:EVS}), Mean Absolute Percentage Error (MAPE)  (\ref{eqn:mape}), Mean Absolute Error (MAE) (\ref{eqn:mae}) and Root Mean Squared Error (RMSE) (\ref{eqn:rmse}).
     \begin{equation}
	   \label{eqn:r2} 
         \textrm{$R^2$} = 1 - \frac{SSE}{SST}  
    \end{equation}
    \begin{equation}
	   \label{eqn:EVS}
         \textrm{EVS} = 1 - \frac{Var(y - \hat{y})}{Var(y)}     
    \end{equation}
     \begin{equation}
	   \label{eqn:mape} 
         \textrm{MAPE} = \frac{1}{n}\sum_{i=1}^n \frac{\mid y_i - \hat{y} \mid}{y_i}        
    \end{equation}
    \begin{equation}
	   \label{eqn:mae} 
         \textrm{MAE} = \frac{1}{n} \sum_{i=1}^n \mid y_i - \hat{y} \mid        
    \end{equation}
     \begin{equation}
	   \label{eqn:rmse} 
         \textrm{RMSE} = \sqrt{\sum_{i=1}^n \frac{(y_i - \hat{y})^2}{n}},        
    \end{equation}
    with $y$ being predicted values and $\hat{y}$ the observed ones with results presented in Table \ref{table:MLresults_TCN}.
     	\begin{table}[htpb]
		\centering
		\renewcommand{\arraystretch}{0.5}
		\caption{Averaged performance metrics for Fig. \ref{fig:ML_results}.}
		\begin{tabular}{|>{\centering\arraybackslash}p{.2\textwidth}|>{\centering\arraybackslash}p{.07\textwidth}|>{\centering\arraybackslash}p{.13\textwidth}|>{\centering\arraybackslash}p{.15\textwidth}|}
			\topline
			\headcol \diagbox{\enskip Metrics}{Scenario} & LoS TCN & NLoS TCN  \\
			\midline
			Explained Variance Score &  0.934 & 0.897  \\
			\midline
			Mean Absolute Percentage Error (\%) & 9.27 &  9.96 \\
			\midline
			Mean Absolute Error & 0.134 &  0.113  \\
			\midline
			Root Mean Squared Error (s) & 0.142 & 0.158  \\
			\midline
			R-Squared ($R^2$) & 0.929 & 0.886  \\       
			\hline
		\end{tabular}
		\label{table:MLresults_TCN}
	\end{table}
          
     	\begin{figure}[htbp]
		\begin{subfigure}{\columnwidth}
			\centerline{\includegraphics[width=0.97\columnwidth, keepaspectratio]{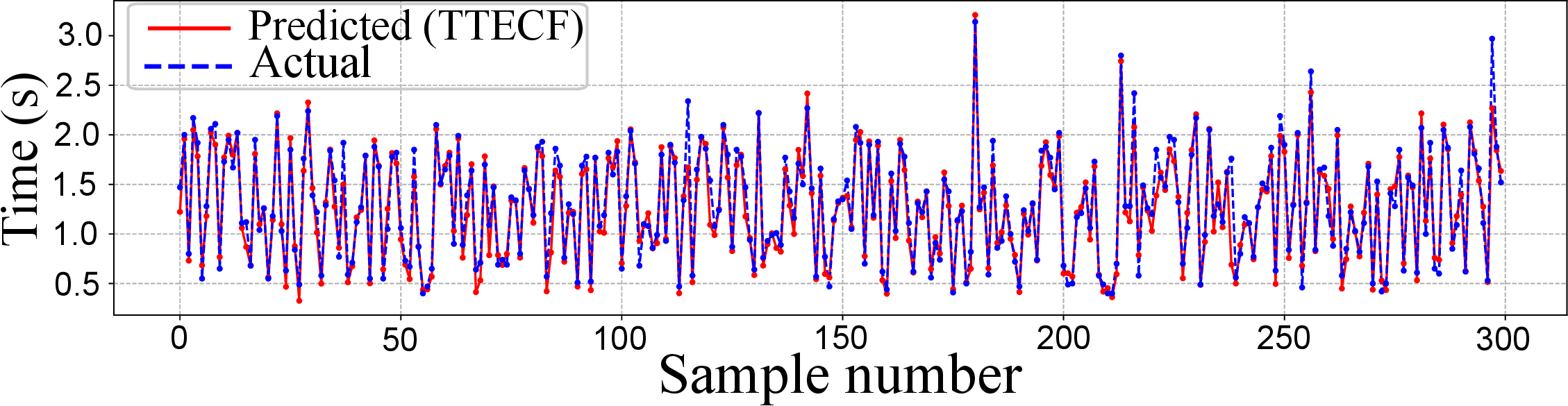}}
			\caption{LoS data from TCN model.}
		\end{subfigure}		
		\begin{subfigure}{\columnwidth}
			\centerline{\includegraphics[width=0.97\columnwidth, keepaspectratio]{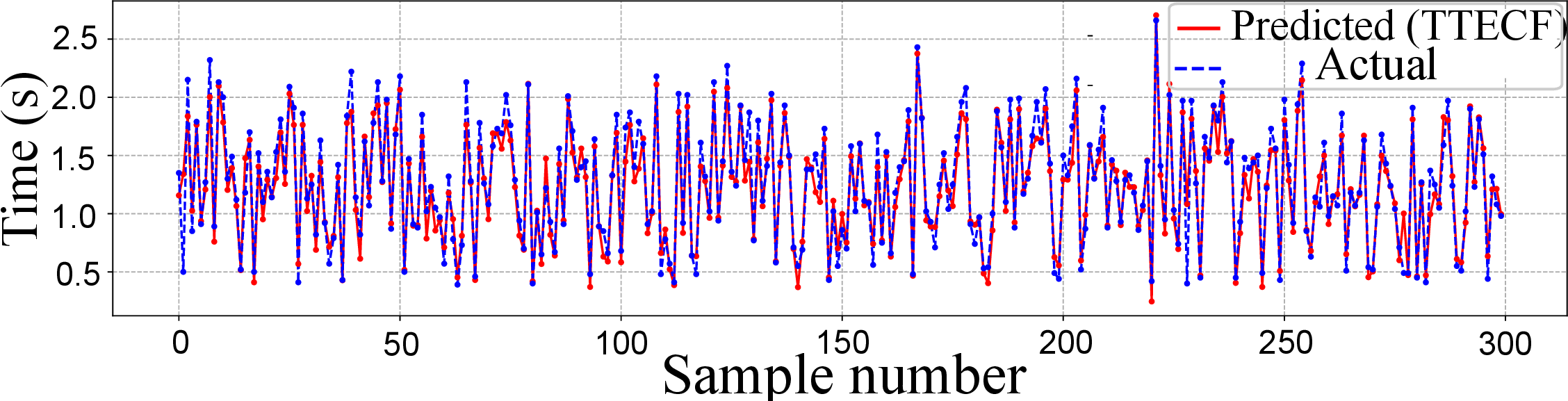}}
			\caption{Non-LoS data from TCN model.}
		\end{subfigure}		
		\caption{Predicted vs. actual time to A3 event entry criterion fulfillment. 5000 epochs, UE velocity 25 m/s and Hysteresis = 1 dBm.}
		\label{fig:ML_results}
	\end{figure}

    The results demonstrate that ML algorithms effectively capture the characteristics and correlations among UE velocity, trajectory, and historical RSRP measurements while integrating the HO parameters with relatively low error rates, as depicted in Fig. \ref{fig:ML_results} (a) and \ref{fig:ML_results} (b) and summarized in Table ~\ref{table:MLresults_TCN}. Notably, while we opted to present results for a single velocity, the TCN framework yields comparable outcomes for both user velocities across accuracy and loss metrics. \newline
    \hspace*{\parindent} The primary goal of this work was not to identify the fastest ML method but to demonstrate the feasibility of the ESHOP scheme, however, we acknowledge that in scenarios requiring extremely low latency, such as millisecond-level handover processes, the computational complexity of TCNs could pose a limitation. For highly time-sensitive applications, a thorough assessment of TCNs' real-time performance is essential. As deploying TCNs in real-time systems might be challenging, the use of fast CPUs, GPUs, or specialized hardware can help avoid significant delays. Additionally, TCNs might be more feasible for systems that make predictions based on shorter data sequences, such as those found in small-cell deployments with high UE velocity, where short but frequent handovers occur. Notably, although not presented in this study, we also explored other machine learning methods, such as Decision Trees \cite{CART}, which can deliver comparable results to TCNs while offering much faster deployment. \newline When faced with challenges in obtaining sufficient and diverse data, the model's behavior might not be reliable in unique scenarios where data is lacking. However, by collecting data across various cell deployment classes and user movement patterns, the model's ability to generalize across different conditions would be enhanced.
    \vspace{-13pt}
    \subsection{ESHOP SCHEME PERFORMANCE}
\begin{figure}[t]
		\centerline{\includegraphics[width=1.1\columnwidth, keepaspectratio]{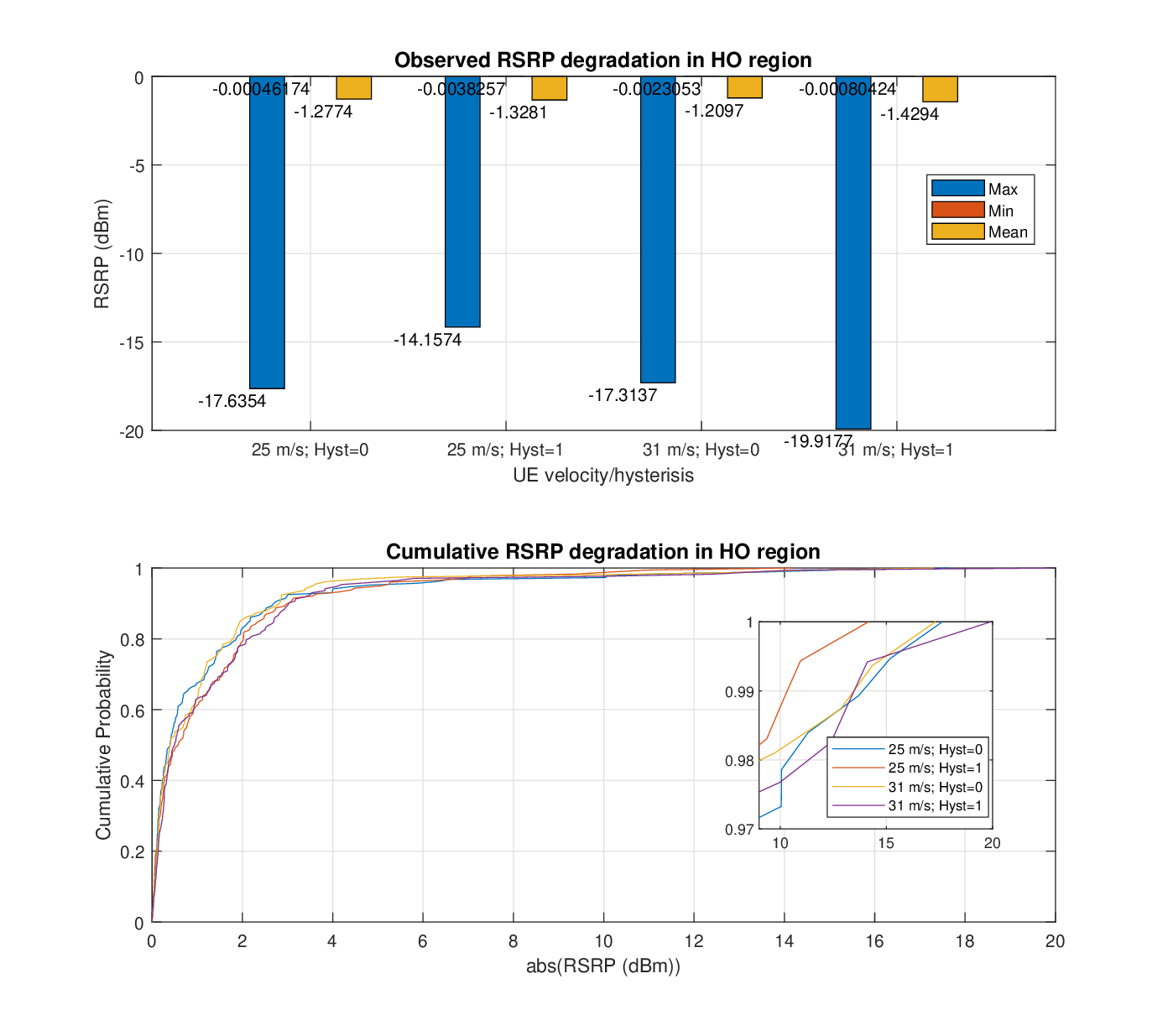}}
		\caption{Serving cell RSRP degradation mitigated by the ESHOP 40 ms after A3 fulfillment.}
		\label{fig:Combined_plot_HO_region_RSRP_drop}
    \end{figure}

    This section evaluates how the ESHOP approach can reduce the likelihood of UE encountering potential PDCCH outage areas at the far end of the HO region by initiating the early parts of the HO preparation stage. Our fingerprint-based approach indirectly incorporates HCP configuration, enabling user context-aware HO optimization. As a result, we refrain from comparing multiple HCP configuration settings or determining the optimality of one approach over another. Instead, our focus is on understanding the behavior and impact of the ESHOP scheme in various scenarios.
    Even without instantly optimized HCP parameters, we can demonstrate the viability of our innovative approach. 
    \begin{figure}[b!]
		\centerline{\includegraphics[width=1.05\columnwidth, keepaspectratio]{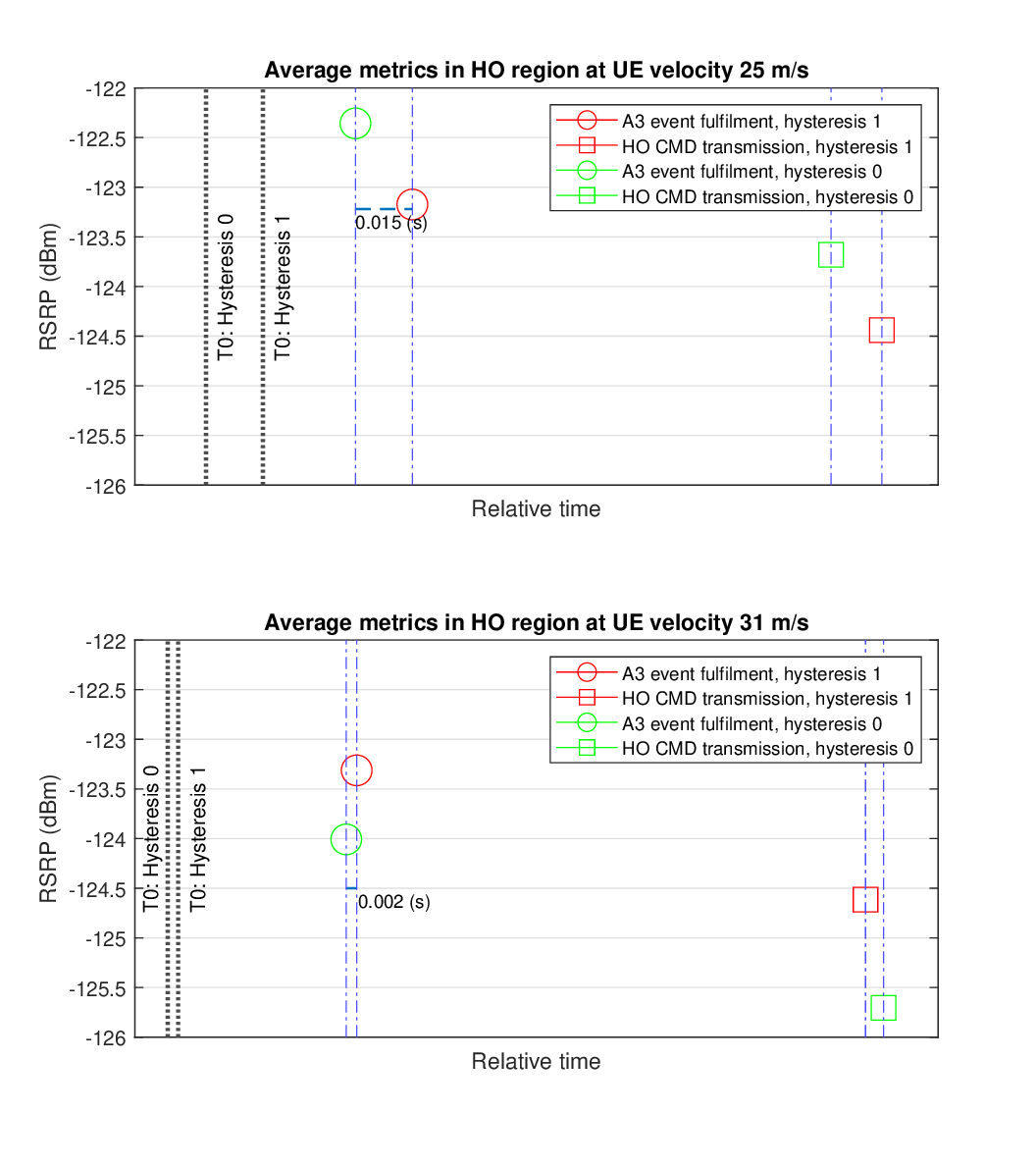}}
		\caption{Expanded handover region and the A3 event dynamics associated with different UE velocities. Due to ML-inferred A3 event \emph{entry criterion fulfillment}, T0 is the new earliest HO trigger instead of the traditional A3 event fulfillment.}
		\label{fig:HO_region}
    \end{figure}
    Due to their relatively high velocity, users enter the HO region where radio conditions may deteriorate significantly, increasing the likelihood of transmission failure for measurement reports or HO commands. This situation is especially likely in denser cell deployments with reduced HO region size. In such mobility scenarios, it is crucial to anticipate impending HOs, as existing HO robustness mechanisms are prone to failure due to their reactive nature, partly caused by the inherent TTT delay. Figure~\ref{fig:HO_region} highlights the importance of user context-aware HO optimization, given that the network lacks control over UE-triggered events. The horizontal dashed lines indicate the average times between \textbf{A3} fulfillments for two different hysteresis settings, which are, as expected, highly dependent on UE velocity. As velocity increases, the time difference between the two events decreases, nearly merging into a single point in time. This observation provides crucial insights into the impact of UE velocity on the distance covered between consecutive UE measurement report intervals. High velocity renders static HCP settings, such as HO hysteresis, ineffective. This underscores the need to advance the start of the handover preparation phase and emphasizes the urgency for context-aware mobility. The value of the ESHOP scheme's ability to pinpoint the \textbf{T0} occasion, regardless of UE velocity, is evident.
    \newline \hspace*{\parindent} The time spent in the HO region, by definition, does not include the TTT duration. In the legacy HO procedure, the TTT interval is often a period of inactivity, during which no signaling occurs until a potential HO event is triggered. However, during this time, signal quality indicators such as the RSRP metric can significantly degrade, potentially leading to RLF if the UE cannot maintain a stable connection with the serving cell long enough to complete the handover to the target cell. As illustrated in Figures \ref{fig:HO_Intra_flowscheme} and \ref{fig:HO_Inter_flowscheme}, the ESHOP scheme effectively anticipates the HO preparation by exactly the TTT duration and integrates HO decision-related signaling into the TTT interval.
    This approach is the cornerstone of the ESHOP scheme as it parallelizes the most time-consuming part of the HO preparation process with the TTT execution. Consequently, the HO command can be sent immediately upon \textbf{A3} event fulfillment, whereas in conventional HO procedures, this would be delayed by HO decision-related signaling. By reducing the overall HO signaling time, ESHOP helps maintain higher RSRP levels, which is particularly crucial in high-speed UE mobility scenarios. 
    Figure  \ref{fig:Combined_plot_HO_region_RSRP_drop} shows the observed channel degradation from \textbf{A3} fulfillment until HO decision signaling is executed, approximately 35-40 ms. Even though the observed RSRP metrics are closely tied to the mobility scenarios modeled in Section \ref{System}, they underscore that rapid UE movement can result in sudden RSRP degradation, causing the signal quality to deteriorate too quickly for the handover process to be completed successfully. 
    By doing so, it helps maintain stable signal quality, preventing the rapid deterioration that could otherwise compromise the handover process. In particular, the extreme RSRP degradation cases depicted in the upper part of Fig. \ref{fig:Combined_plot_HO_region_RSRP_drop}, with a cumulative probability close to the 98th percentile, would likely result in RLF and HOF. Maintaining an adequate RSRP level is a primary and direct benefit of ESHOP's ability to parallelize HO-related signaling procedures.
    \newline \hspace*{\parindent} We highlight that the mobility scenarios modeled in the system described in Section \ref{System} concentrated on intra-gNodeB mobility and did not include HO decision-related signaling. To assess the performance of the ESHOP scheme, we used measurements from a commercial-grade 5G test network as a reference.
    The measured signaling times averaged between between 15-35 ms for legacy procedures as shown in steps 2-4, Fig.  \ref{fig:HO_Intra_flowscheme} and steps 2-5, Fig.  \ref{fig:HO_Inter_flowscheme} which fits entirely within the shortest specified TTT duration of 40 ms.

    \section{CONCLUSIONS} 
    This study demonstrates how the ESHOP scheme mitigates serving cell RSRP degradation within the HO region by initiating the HO preparation phase earlier. We focus on simple deployments and mobility scenarios to highlight the novel approach presented, specifically targeting small-cell deployments due to their frequent HOs, which are not as prevalent in large-cell networks. The dense, small-cell model results in rapid and frequent HOs, providing a robust test of the proposed algorithm. In contrast, larger network deployments would complicate the analysis due to their complexity and the challenges in generating, collecting, and processing measurement data, exceeding the scope of this study. \newline \hspace*{\parindent} The proposed ESHOP scheme reduces or removes the forced inactivity by parallelizing the HO preparation and the TTT intervals, and reduces instances of severe link degradation and HO failures. Hence, a user experiences more stable connections and increases the 5G NR network capacity. Allthough our study focuses on A3 events, the deployed ML model can be trained on any predefined HO event due to its flexible implementation. The proposed ESHOP scheme is designed for the network side, however, its predictive capabilities for upcoming handover events can beneficially be integrated on the UE side, enhancing techniques such as Conditional HO. \newline \hspace*{\parindent} Future studies should combine narrow-beam Channel State Information Reference Signals (L1-RSRP) with wide-beam SSB (L3-RSRP) measurements at the cell edge to enhance data resolution by capturing both levels of beam measurement information. We encourage future handover optimization studies to implement multiple instances of the ESHOP scheme across various network nodes, allowing them to share insights related to load balancing and handover coordination. Protocols like Xn facilitate communication between base stations, enabling these nodes to optimize both user experience and network efficiency. This approach would promote a more comprehensive and effective handover optimization strategy across the network. \\
    \hspace*{\parindent} Additionally, we believe that attention-enabled generative AI models could significantly enhance the ESHOP framework. Attention models, particularly Transformer-based architectures, generally outperform TCNs for tasks that involve capturing long-range dependencies, which are likely to occur in more complex network environments. Their ability to model long-term dependencies and correlations makes Transformer-based models ideal for time series applications, such as forecasting. Since the proposed ESHOP model relies on the dataset's capacity to capture key fingerprinted features, such as a user's trajectory and velocity, incorporating a Transformer-based model for trajectory analysis could potentially improve the accuracy of the ESHOP framework. Although deploying Transformer-based models in commercial systems may pose challenges, GPUs or specialized hardware can help mitigate delays, particularly in complex commercial network environments.\\

\section*{REFERENCES}

\def\refname{\vadjust{\vspace*{-1em}}} 

\vspace{-12pt}
\begin{IEEEbiography}[{\includegraphics[width=1in,height=1.25in,clip,keepaspectratio]{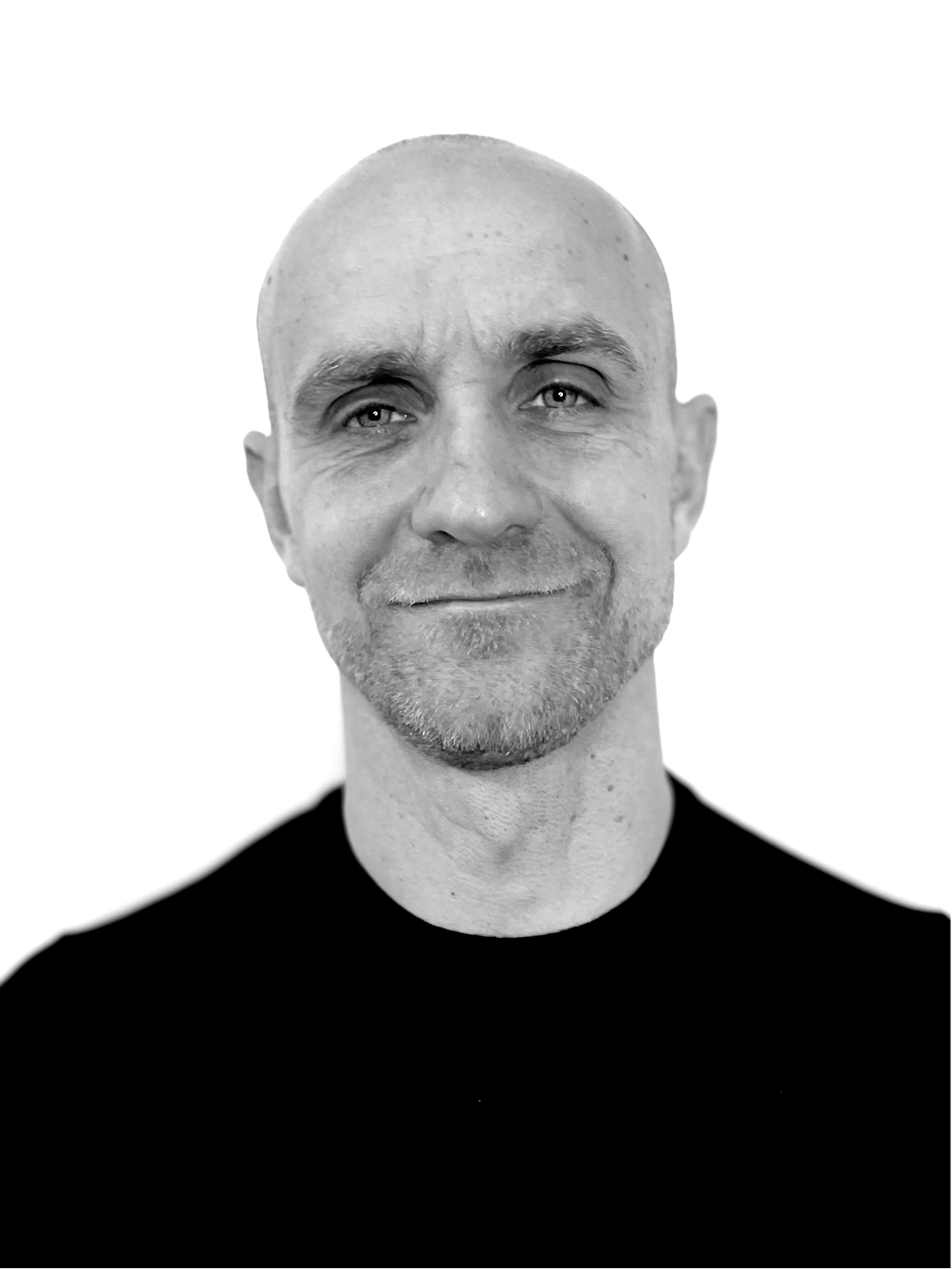}}]
{Dino Pjanić}~(Student Member, IEEE), received an M.S.E.E. degree with an emphasis on wireless telecommunications from the Blekinge Institute of Technology (BTH), Karlskrona, Sweden, in 2005. \newline 
\indent He is currently employed at Ericsson where he pursues industrial PhD studies in electrical engineering at Lund University, Sweden. His research interests include optimization and applications of machine learning for efficient massive MIMO operation in the radio access domain of cellular networks.
\end{IEEEbiography}
\vspace{-13pt}
\begin{IEEEbiography}[{\includegraphics[width=1in,height=1.25in,clip,keepaspectratio]{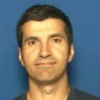}}]{Alexandros Sopasakis}~(Member, IEEE),
received the B.S.,
M.S. and Ph.D. degrees in applied mathematics
all from Texas A\&M University at College Station, in 1991, 1993 and 2000 respectively.
He held associate scientist positions in mathematics at Georgia Tech. in Atlanta, Univ. of California at Berkeley and Courant Institute, NYU. His current research interests include hybrid systems consisting of stochastic dynamics coupled to machine learning models which are reinforced for learning with data. Applications range from solving SDEs to modeling and forecasting traffic network evolution from camera images.
\end{IEEEbiography}
\vspace{-13pt}
\begin{IEEEbiography}[{\includegraphics[width=1in,height=1.25in,clip,keepaspectratio]{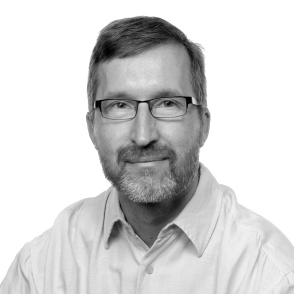}}]
{Andres Reial}~(Senior Member, IEEE), received a Ph.D. degree in Electrical Engineering from the University of Virginia in 2000. He is currently working at Ericsson Research, Lund, Sweden. His research interests include AI solutions for physical layer processing and practical deployment aspects of massive- and distributed-MIMO systems. 
\end{IEEEbiography}
\vspace{-13pt}
\begin{IEEEbiography}[{\includegraphics[width=1in,height=1.25in,clip,keepaspectratio]{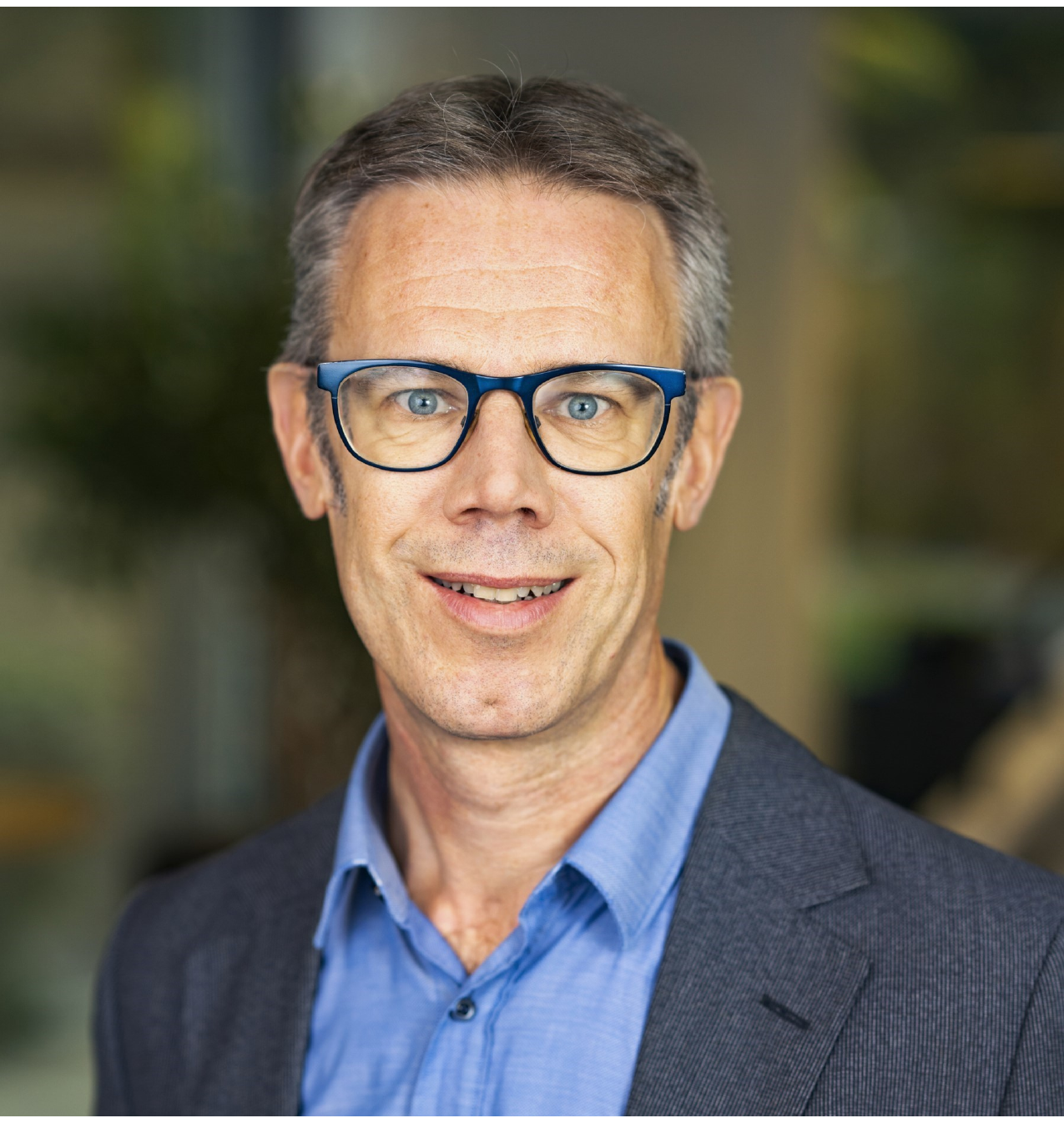}}]
{Fredrik Tufvesson}~(Fellow, IEEE), received his Ph.D. in 2000 from Lund University in Sweden. After two years at a startup company, he joined the department of Electrical and Information Technology at Lund University, where he is now a professor of radio systems. His main research interest is the interplay between the radio channel and the rest of the communication system with various applications in 5G/6G systems such as massive MIMO, distributed MIMO, mm wave communication, vehicular communication and radio-based positioning.
Fredrik has authored around 120 journal papers and 200 conference papers, he is a fellow of the IEEE and his research has been awarded the Neal Shepherd Memorial Award (2015) for the best propagation paper in IEEE Transactions on Vehicular Technology, the IEEE Communications Society best tutorial paper award (2018, 2021) and the IEEE Signal Processing Society Donald G. Fink overview paper award 2023.
\end{IEEEbiography}

\vfill\pagebreak
\end{document}